%% file: main.tex
\documentclass[10pt,twocolumn,letterpaper]{article}

\usepackage[pagenumbers]{wacv} 

\input{preamble}

\definecolor{wacvblue}{rgb}{0.21,0.49,0.74}

\def\wacvPaperID{1773} 
\def\confName{WACV}
\def\confYear{2027}

\renewcommand{\paragraph}[1]{%
    \par\medskip
    \noindent\textbf{#1}\quad
}
\newcommand\oursname{\text{Alignment-Aware Bridge Matching}\xspace}
\newcommand\ours{\text{A²BM}\xspace}

\title{\ours: \oursname for Image-to-Image Translation}

\author{
Aimi Okabayashi$^\dagger$\textsuperscript{1},
Georges Le Bellier$^\dagger$\textsuperscript{2,4},
Nicolas Audebert\textsuperscript{3},\\
Charlotte Pelletier\textsuperscript{1},
Thomas Corpetti\textsuperscript{5},
Nicolas Courty\textsuperscript{1}\\[0.6em]
\textsuperscript{1} Université Bretagne Sud, IRISA, UMR CNRS 6074, F-56000 Vannes, France\\
\textsuperscript{2} Inria, ENS de Lyon, CNRS, Université Claude Bernard Lyon 1, LIP, UMR 5668, 69342 Lyon, France\\
\textsuperscript{3} Université Gustave Eiffel, ENSG, IGN, LASTIG, F-94160 Saint-Mandé, France\\
\textsuperscript{4} Conservatoire national des arts et métiers, CEDRIC, F-75141 Paris, France\\
\textsuperscript{5} CNRS, UMR 6554 LETG, Univ. Rennes 2, 35043 Rennes, France
\\[0.2em]
{\tt\small
aimi.okabayashi@univ-ubs.fr,
georges.le-bellier@inria.fr
}
}

\begin{document}
\maketitle

\input{sec/0_abstract}

\noindent{\small $^\dagger$ These authors contributed equally to this work.}

\input{sec/1_intro}

\input{sec/2_related}

\input{sec/3_method}

\input{sec/4_experiments}
\input{sec/5_results}
\input{sec/6_conclusion}

\section*{Acknowledgement}

This work was conducted as part of the research projects \href{https://mage.science/}{\textsc{MAGE}} (ANR-22-CE23-0010) and \href{https://moni-tree.github.io/}{\textsc{MONI-TREE}} (ANR-23-CE04-0017) funded by the \emph{Agence
Nationale de la Recherche}. This work was also supported by Collège doctoral de Bretagne. This work was granted access to the HPC resources of IDRIS under the allocations AD011014327R2 and AD011015868R1 made by GENCI.

{
    \small
    \bibliographystyle{ieeenat_fullname}
    \bibliography{main}
}

\newpage

\input{supplemental}
\end{document}

%% file: preamble.tex
\usepackage{graphicx}
\usepackage{booktabs}
\usepackage{algorithm}
\usepackage{algorithmic}
\usepackage{graphbox}
\usepackage{float}
\usepackage[detect-all,per-mode=symbol]{siunitx}
\DeclareSIUnit\pixel{px}
\usepackage[table]{xcolor}
\definecolor{customblue}{RGB}{166,200,254}
\usepackage[accsupp]{axessibility}  
\usepackage{orcidlink}
\usepackage{comment}
\usepackage{subcaption}
\usepackage{tikz}
\usepackage{caption}
\usepackage{xspace}
\usepackage{amsmath}
\usepackage{amssymb}

\renewcommand{\eg}{\textit{e.g.}\xspace}
\renewcommand{\ie}{\textit{i.e.}\xspace}

\renewcommand{\etal}{\textit{et al.}\xspace}



%% file: sec/0_abstract.tex
\begin{abstract}
Paired image-to-image translation underpins a wide range of computer vision tasks, including image editing, sensor translation, and domain adaptation. Bridge matching and flow matching have recently emerged as powerful frameworks, extending diffusion models to arbitrary source and target distributions. However, their standard formulations assume perfectly aligned training pairs, treating all source–target correspondences as equally reliable. In practice, real-world applications often involve \emph{weakly aligned} pairs due to changes of acquisition conditions, including {\em e.g.} asynchronous captures, different illuminations, or misregistration. In this work, we introduce Alignment-Aware Bridge Matching  (\ours), a bridge matching method that leverages image pairs alignment during training. By incorporating alignment scores, the model learns to disentangle true semantic correspondences from misalignment artifacts. At inference time, we use the alignment score as a control variable over translation fidelity, with strongly aligned outputs obtained when prompting the model with the highest alignment score. We validate \ours on both controlled synthetic experiments and on challenging real-world tasks, including cross-sensor super-resolution and pixel-space unsupervised domain adaptation. In all settings, \ours consistently improves translation fidelity over strong GAN-, diffusion-, and Schrödinger bridge-based baselines, establishing alignment conditioning as a principled solution for image translation models with \emph{weakly aligned} data.

\end{abstract}

%% file: sec/1_intro.tex
\section{Introduction}
\label{sec:intro}

\begin{figure}[h]
\begin{center}
\centerline{\includegraphics[width=0.94\columnwidth, trim=0 10 0 0]{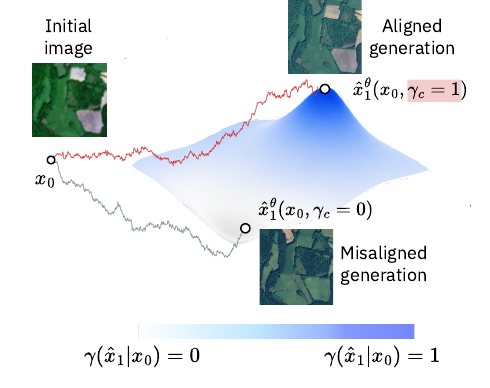}}
\caption{\textbf{Alignment-Aware Bridge Matching (\ours)} performs image translation conditioned on an \textit{alignment score} $\gamma_c$. At inference time, generation is guided by a user-specified alignment score, with $\gamma_c=1$ enforcing the strongest alignment to the source image $x_0$, i.e. highest $\gamma(\hat{x}_1|x_0)$, and thus the highest source fidelity during translation.}
\label{fig:teaser}
\end{center}
\vspace{-12mm}
\end{figure}
Image-to-image translation aims to learn a mapping that transforms an image from a source domain to a target domain while preserving source content information. It encompasses a large number of applications in image editing (e.g. inpainting, super-resolution \cite{zhang_multiple_2020}, dehazing \cite{Qu_2019_CVPR} or relighting \cite{yang_relighting_2025}), perception (e.g. monocular depth estimation \cite{pilzer_unsupervised_2018}) and cross-modal generation (e.g. RGB-to-thermal \cite{xiao2025thermalgen} and radar-to-optical translation~\cite{bai_conditional_2024}). In particular, it is a staple of domain adaptation for computer vision \cite{murez2018image, pizzati2020domain}.

Image translation literature has largely focused on two extreme regimes: \emph{perfectly paired} or \emph{fully unpaired} data. In the perfectly paired setting, each source image is associated with a target image depicting the same scene, enabling direct supervision. Recent generative paradigms, diffusion, flow matching, and bridge matching perform high quality translation in this setting. In contrast, \emph{unpaired} methods remove the need for explicit correspondences, and rely on cycle consistency \cite{cyclegan_zhu} or optimal transport \cite{liu_rectified_2022, shi_diffusion_2023, bortoli_schrodinger_2024} to learn mappings without explicit pairs. While attractive from a data collection perspective, these approaches struggle to preserve fine-grained source content during translation.

Many real-world applications fall between these two extremes. We refer to this setting as \emph{weakly aligned} image translation, where source and target images are paired but imperfectly matched. Such pairs are frequently affected by geometric shifts (e.g., camera motion or parallax), content changes from desynchronized captures, or acquisition noise, all of which introduce misalignment. These problems are common in medical imaging and remote sensing, where cross-sensor translation (e.g., SAR to optical \cite{bai_conditional_2024,yang2022sar}, MRI T1 to FLAIR \cite{yang_mri_2020}) must nonetheless preserve semantic content, since the outputs support quantitative downstream analysis such as tumor segmentation  \cite{yang_mri_2020} or land-cover mapping \cite{le2026floweo}. Satellite imagery especially suffers from misalignments due to desynchronized acquisitions. The larger the time gap between two images of the same place, the more changes. Other factors can introduce mismatches, \eg occlusions due to clouds, or differences in sensor orbit and orientation making the pairs not accurately co-registered. Perfectly aligned pairs are thus extremely difficult to collect in remote sensing, making misalignment a fundamental obstacle to accurate image translation.

Despite its practical importance, weakly aligned image translation remains relatively underexplored. Recent flow and bridge matching formulations implicitly assume perfectly aligned training pairs. On \textit{weakly aligned} data, mismatched correspondences introduce corrupted supervision, leading to degraded content preservation during translation. Existing approaches for the \textit{weakly paired} setting adapt methods developed for unpaired translation to handle imperfect correspondences.

This work introduces \textbf{Alignment-Aware Bridge Matching} (\ours), a bridge matching framework for weakly aligned image translation (\cref{fig:training}). We explicitly model alignment of image pairs and use it as a conditioning signal during training, enabling the model to separate relevant features from misaligned ones. To this end, we propose two alignment scores: a lightweight metadata-based score designed to leverage prior knowledge when available, and a feature-space similarity score based on pretrained visual encoder embeddings when no domain knowledge is available. At inference time, we use the alignment score as a control variable over translation fidelity, with high fidelity images obtained by prompting the model with highest alignment score.
Our contributions are as follows:

\begin{enumerate}[itemsep=1em]
    \item We introduce \ours, an alignment-aware bridge matching that enables robust image-to-image translation from weakly paired data.
    \item We validate the approach in a controlled setting by introducing synthetic misalignments into a perfectly paired dataset.
    \item We demonstrate superior translation fidelity over strong baselines across real-world translation tasks in remote sensing: post-/pre- flood image translation, historical image translation, and cross-sensor super-resolution. 
\end{enumerate}

%% file: sec/2_related.tex
\section{Related Work}
\label{sec:related}

\begin{figure*}[t]
\begin{center}
\centerline{\includegraphics[width=\textwidth, trim=0 0px 0px 0, clip]{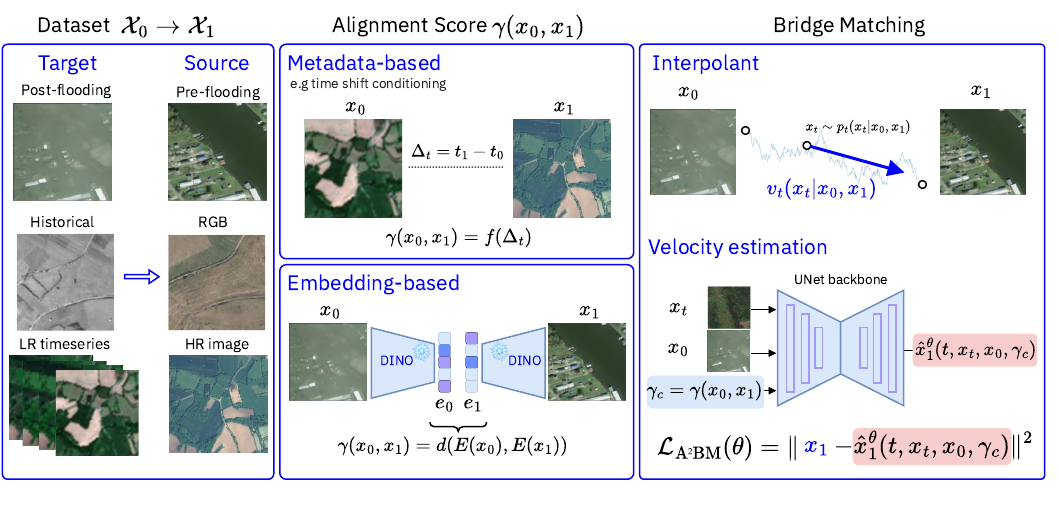}}
\vspace{-2mm}
\caption{\textbf{\ours~overview.} \ours is a \emph{weakly aligned} image translation framework, covering challenging tasks such as pre/post-disaster domain adaptation and cross-sensor super-resolution. It leverages an alignment score $\gamma$ to learn a transport robust to mismatches in training pairs. This scores can either be derived from available metadata or from the cosine similarity between pretrained visual encoder embeddings, and injected into the UNet backbone as a conditioning signal.}
\label{fig:training}
\end{center}
\vspace{-10mm}
\end{figure*}

\paragraph{Image-to-image translation} Image-to-image translation aims to learn a mapping between source and target distributions while preserving source content. The seminal work of Pix2Pix \cite{pix2pix} established the standard paradigm for paired image translation using conditional GANs. Since then, image-to-image translation has been applied to many tasks, including sensor translation \cite{yang2022sar}, super-resolution \cite{esrgan}, dehazing \cite{shao2020domain}, monocular depth estimation \cite{aleotti2018generative}, and domain adaptation \cite{murez2018image}.
The strong generative capabilities of diffusion models (DMs) \cite{song_score-based_2020, rombach2022, esser2024scaling} have motivated their use for image translation by conditioning the generation on a source image \cite{saharia_palette}. However, these models fundamentally rely on conditioning mechanisms to guide the generation process from noise to image, limiting the faithfulness of the resulting translations to the source image.

Building on this, bridge matching \cite{zhou_denoising_2023,bortoli_augmented_2023} and flow matching \cite{albergo_stochastic_2023, lipman_flow_2022, peluchetti2023non} have generalized diffusion processes by defining probability paths between arbitrary distributions. These new paradigms have been successfully applied in image translation tasks \cite{liu_i2sb_2023, chadebec_lbm_2025}, including depth estimation \cite{gui2025depthfm}, sensor translation \cite{xiao2025thermalgen} and domain adaptation \cite{le2026floweo}. 
Considerable effort has been devoted to extending bridge matching beyond perfectly paired datasets, but existing approaches either assume exact correspondences \cite{albergo_stochastic_2024, bortoli_augmented_2023, liu_i2sb_2023, somnath_aligned_2023} or rely on optimal transport couplings to handle unpaired settings \cite{tong_simulation-free_2023, de2021diffusion, shi_diffusion_2023, bortoli_schrodinger_2024}, leaving the intermediate, weakly aligned regime largely unaddressed.

\paragraph{Image-to-image translation with weakly aligned data} Weakly aligned image pairs arise naturally in many real-world settings where obtaining pixel-perfect correspondence is expensive or infeasible. The nature of misalignment varies by task, and ad hoc strategies were developed to correct some aspects of misalignment during training. For example, RegGAN \cite{reggan_kong} adds a deformation field registration network to Pix2Pix \cite{pix2pix} to correct spatial misalignment dynamically. Other works build upon CycleGAN, a model originally designed for unpaired images \cite{cyclegan_zhu} using two GANs to enforce a cycle-consistency loss, ensuring that an image translated to a target domain can be mapped back to its original form. For example, Xia \etal \cite{xia2023coarsely} adapt it to weakly aligned settings by adding different losses to the CycleGAN training, such as masking foreground objects to focus the model on relevant objects. Other works adopt semi-supervised approaches, leveraging a small amount of paired data guiding larger collections of unpaired samples \cite{theodoropoulos_feedback_2024}.

 \paragraph{Learning generative models with imperfect data} 
Training data quality is a critical factor in learning high-quality generative models \cite{schuhmann2022laion, penedo2024fineweb}, often motivating curation and preprocessing pipelines that filter out low-quality or incorrectly annotated samples. However, there has been growing interest in effectively using all available data, so that even low-quality samples contribute positively to the training. \textit{Ambient diffusion} \cite{daras2023ambient} in particular proposes training diffusion models by exploiting low-quality data only at high noise levels, where the distinction between high- and low-quality samples becomes negligible. These approaches have demonstrated effectiveness across diverse applications, including image and protein generation \cite{daras2023ambient, daras2025ambient, daras2025ambient_protein}. Similarly, Dufour \etal \cite{dufour2024dont} argue against discarding poorly annotated images during text-to-image diffusion models training. Instead, by conditioning the backbone on the accuracy of the image caption, their coherence-aware diffusion models generated images with higher fidelity with respect to the text prompt. In image translation, however, recent bridge matching works primarily use perfectly paired data \cite{liu_i2sb_2023, chadebec_lbm_2025, gui2025depthfm}, and do not address the weakly aligned setting, which is precisely the gap that \ours is designed to fill.

%% file: sec/3_method.tex
\section{Alignment-Aware Bridge Matching}
\label{sec:method}

This section introduces \oursname (\ours), our time-dependent transport between two image distributions based on bridge matching \cite{zhou_denoising_2023, albergo_stochastic_2023} tailored for \textit{paired} datasets with \textit{alignment} mismatches. In particular, we propose a new image translation conditioned on an \textit{alignment score} $\gamma$, either extracted from metadata or derived from pretrained vision encoders' embeddings.
The alignment score $\gamma(x_0, x_1)$ is computed between image pairs $(x_0, x_1)$ during training.
At inference time, prompting the translation process with different $\gamma$ values allow us to control the adherence of the output to the source image, reaching higher translation fidelity when $\gamma$ increases.

\subsection{Bridge Matching}

Bridge matching extends diffusion models by modeling the transport between an initial distribution $p_0$ and a target distribution $p_1$ with a stochastic differential equation that defines a probabilistic bridge between the two:

\begin{equation}
    dX_t = v(t, X_t)dt + \sigma dW_t, \quad X_0 \sim p_0, X_1 \sim p_1
\label{eq:diffusion_bridge}
\end{equation}

where $\sigma dW_t$ is a scaled Brownian motion and $v(t, \cdot) : [0, 1] \times \mathbb{R}^d \rightarrow \mathbb{R}^d$ is the \textit{drift} function, guiding images from $p_0$ to $p_1$, that needs to be learned. To learn such a drift function, we first sample an image pair $(x_0, x_1)$ from the joint distribution, also called \textit{coupling}, $p(x_0, x_1)$. We use \textit{data-dependent couplings} \cite{albergo_stochastic_2024} $p(x_0, x_1) = p(x_1\mid x_0)p(x_0)$ that allow to use the images from \textit{paired} datasets:
\begin{equation}
    (x_0, x_1) \sim p(x_1\mid x_0)p(x_0).
\label{eq:data-dependent_coupling}
\end{equation}

Then, we compute an intermediate $x_t$ from the marginal distribution $p_t(x_t\mid x_0, x_1)$ using a \textit{stochastic interpolant}:

\begin{equation}
    x_t = (1-t)x_0 + tx_1 + \sigma\sqrt{t(1-t)}z, \quad z\sim\mathcal{N}(0, I_d).
\label{eq:interpolant}
\end{equation}

\textit{Augmented bridge matching} (AugBM) \cite{bortoli_augmented_2023} conditions the drift function on the source image $x_0$ to preserve the training coupling. In fact, although standard bridge matching models are trained on paired samples $(x_0, x_1)$, the learned sampling process does not recover the training correspondence. Given a train input $x_0$, the generated image $x_1^{\text{gen}}$ often differs from its paired target $x_1$. AugBM adopts the following loss:

\begin{equation}
    \mathcal{L}_\text{AugBM}(\theta) = \mathbb{E}_{\substack{(x_0, x_1) \sim p(x_0, x_1) \\ x_t \sim p_t(x_t\mid x_0, x_1) \\ t \sim \mathcal{U}([0, 1])}} \Bigg[\Big\lVert v_\theta(t, x_t, x_0) - \frac{x_1 - x_t}{1 - t} \Big\rVert^2\Bigg].
\label{eq:abm_loss}
\end{equation}

Once the drift estimator is trained, we numerically solve the SDE \cref{eq:diffusion_bridge} from $t=0$ to $t=1$ to translate the initial image $x_0$ into a generated image $x_1^{\text{gen}}$.

\subsection{\oursname}

AugBM is designed for image translation on \textit{paired} datasets. Yet, in real-world image pairs collected are often imperfect. Acquisition conditions, sensor noise, and temporal shifts can introduce semantic changes or geometric shifts that reduce the alignment between $x_0$ and $x_1$. Crucially, AugBM is unaware of the varying degrees of misalignment existing between image pairs and treats all training pairs as equally aligned, leading to lower fidelity translations.

\paragraph{\oursname} \ours overcomes this limitation by leveraging \textit{alignment scores} to capture the different levels of alignment among training pairs and better preserve source information at inference. We assume that we can compute an alignment score $0 \leq \gamma(x_0, x_1) \leq 1$ for each image pair $(x_0, x_1)$ of the training set.
Images that are geometrically aligned and share the same semantics receive higher alignment scores, whereas dissimilar images yield scores close to zero.
We detail how to build such a score for various applications in \cref{sec:alignment_score_design}.

Then, we enrich the definition of 
$p(x_1|x_0)$ with the alignment score as a latent variable. This decomposition captures the existence of multiple possible final images $x_1$ for a given initial image $x_0$, depending on the alignment score $\gamma$:
\begin{equation}
    p(x_1|x_0)
    = \int p(x_1|x_0, \gamma)\, p(\gamma)\, \text{d}\gamma\text{,}
\end{equation}

where $p(\gamma)$ denotes the marginal distribution of alignment scores induced by the dataset. The transition of data points from $x_0$ to $x_1$, conditioned on an alignment score $\gamma$, is driven by a $\gamma$-dependent drift given by (mathematical derivation in Appendix E):
\begin{equation}
\resizebox{\columnwidth}{!}{$
v_t(x_t \mid x_0, \gamma) =  \mathbb{E}_{p(x_1 \mid x_t, x_0, \gamma)}\left[ \frac{x_1 - x_t}{1-t} \right] = \frac{\mathbb{E}_{p(x_1 \mid x_t, x_0, \gamma)}[x_1] - x_t}{1-t}.
$}
\end{equation}

In what follows, we distinguish the alignment score $\gamma(x_0, x_1)$, computed during training, from the conditioning signal $\gamma_c$ supplied to the neural network.

\paragraph{Training} During training (\cref{fig:training}), we sample a pair $(x_0, x_1)$ from the data-dependent coupling, then compute its alignment score $\gamma(x_0, x_1)$. We compute an intermediate $x_t$ with the interpolant \cref{eq:interpolant}. Accordingly, the network is optimized to predict the target $x_1$ from $x_t$, conditioned on $x_0$ and $\gamma_c=\gamma(x_0, x_1)$, using a standard regression loss:

\begin{equation}
\resizebox{0.95\columnwidth}{!}{$
\mathcal{L}_\text{\ours}(\theta)=
\mathbb{E}_{\substack{(x_0, x_1) \sim p(x_1, x_0) \\ x_t \sim p_t(x_t\mid x_0, x_1) \\ t \sim \mathcal{U}([0, 1])}}
\Bigg[\Big\lVert \hat{x}_1^\theta\big(t, x_t, x_0, \gamma_c\big) - x_1 \Big\rVert^2\Bigg].
$}
\end{equation}

\paragraph{Inference} We want to follow the transport defined by the best-aligned training pairs to achieve the highest image translation fidelity (\cref{fig:teaser}). At inference time, generate new images $\hat{x}_1^\theta(x_0, \gamma_c)$ from $x_0$ by numerically solving the bridge SDE (\cref{eq:diffusion_bridge}) with the drift $\frac{\hat{x}_1^\theta(t,x_t, x_0, \gamma_c)-x_t}{1-t}$.

\subsection{Alignment score design}
\label{sec:alignment_score_design}

The final part of \ours consists in designing a meaningful alignment score $\gamma$. Our goal is for this score to capture the quality of an image pair, {\em i.e.} how much they correspond despite natural mismatches that can occur. We follow a general principle: we leverage domain-specific prior knowledge when it is available, and rely on pretrained embeddings otherwise that do not require any domain knowledge or manual annotation. This leads to two scores: a lightweight metadata-based score and a semantic similarity based on a pretrained model feature space.

\paragraph{Alignment score from metadata} Metadata provide highly informative cues regarding how well two images are aligned, especially in Earth observation, for which spatio-temporal metadata is commonly available. Satellite image pairs are generally acquired over the same geographical area, but at different times, sometimes with different sensors. As the time gap increases, more changes occur (cloud coverage, seasonal variation of vegetation, constructions, etc.). As a first approximation, the discrepancy between pairs can be considered as proportional to this time gap. Therefore, we define a temporal alignment score by rescaling the temporal gaps $\Delta t$ into a normalized score $\gamma \in [0,1]$:
\begin{equation}
    \gamma(x_0, x_1) = 1- \Delta t(x_0, x_1)/\Delta t_{\text{max}}.
    \label{eq: temporal_score}
\end{equation}

\paragraph{Alignment score based on embeddings}
\label{subsec:align_emb}
In the general case, we cannot assume that metadata will be available. To compute a meaningful alignment score when no prior knowledge is available, we rely on rich semantic features to capture image similarity. We leverage pre-trained visual encoders embeddings to project images in a feature space where distances reflect visual similarity. Let $E$ denote a pretrained visual encoder that maps data points $x$ to embeddings, \textit{i.e.}, $e = E(x)$. We define the alignment score $\gamma$ from the Euclidean distance on the embeddings. As the distances are often tightly concentrated around a single value, directly conditioning on them can hinder the model's ability to discriminate between different levels of alignment. To mitigate this, we discretize the $\gamma$ values into $k$ uniformly distributed bins scaled within $[0, 1]$:

\begin{equation}
\gamma(x_0, x_1) = 1-B(\lVert E(x_0) - E(x_1) \rVert),
\label{eq:similarity_score}
\end{equation}
where $B: \mathbb{R} \to \{\gamma_1, \dots, \gamma_k\}$ maps the distance to its corresponding bin (more details on bins in Appendix B.2).

%% file: sec/4_experiments.tex
\section{Data and experimental settings}
\label{sec:experiments}

To evaluate the effectiveness of the alignment score conditioning, we first design a synthetic experiment simulating super-resolution scenarios with varying levels of misalignment. We then assess \ours on three real-world image translation tasks: cross-satellite super-resolution, historical to modern domain adaptation of aerial images, and post-/pre-flood domain adaptation of satellite images.

\subsection{Controlled experiments on synthetic data}

\begin{figure}[tbp]
    \centering
    \begin{subfigure}[b]{0.9\linewidth}
        \centering
        \includegraphics[width=\linewidth]{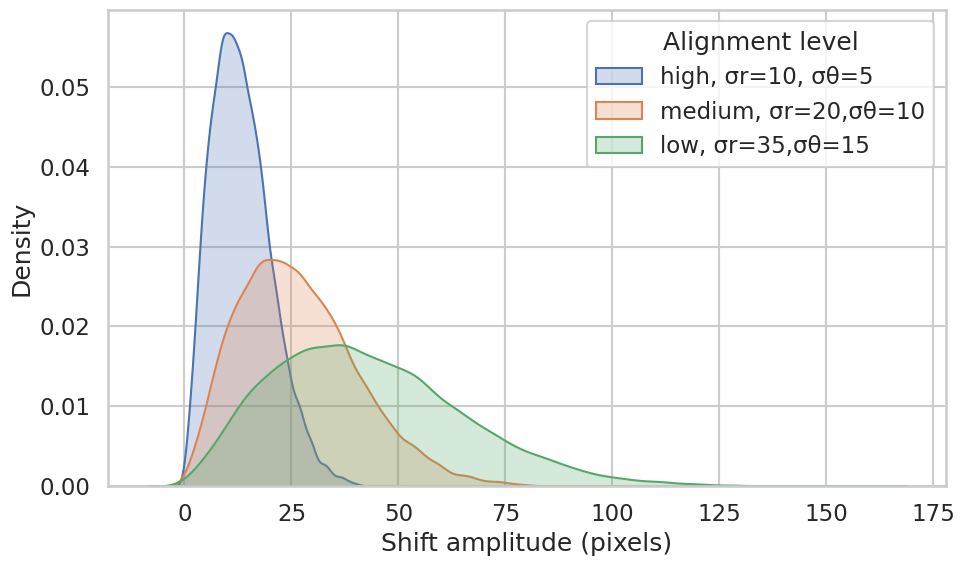}
        \label{fig:div2k_alignment}
    \end{subfigure}

    \vspace{-0.30cm}

    \begin{subfigure}[b]{\linewidth}
        \centering
        \includegraphics[width=\linewidth]{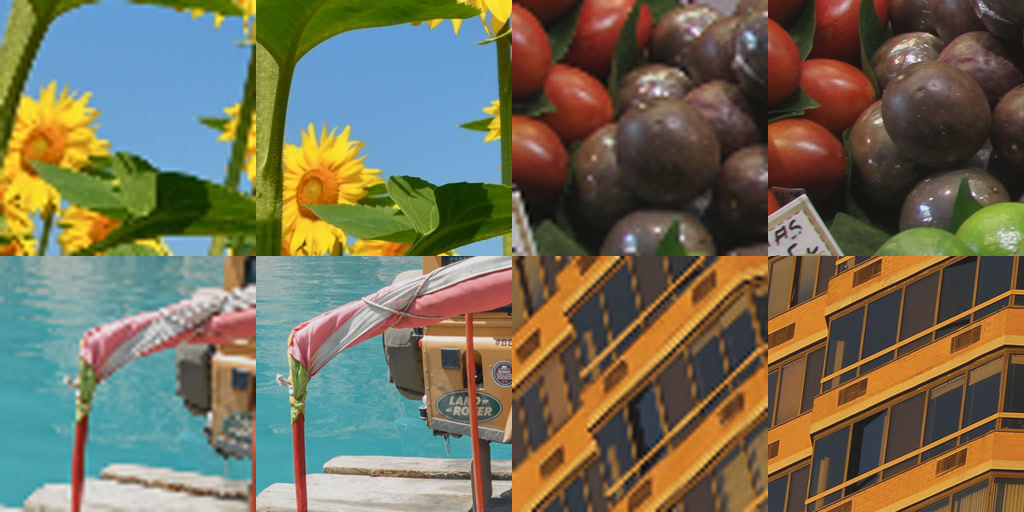}
        \label{fig:div2k_examples}
    \end{subfigure}
    \vspace{-10mm}
    \caption{DIV2K synthetically misaligned datasets. (Top) Distribution of spatial shifts applied to DIV2K \cite{div2k_Agustsson_2017_CVPR_Workshops} divided into: \emph{High}, \emph{Medium}, and \emph{Low} alignment levels. (Bottom) Samples from the synthetically misaligned datasets.}
    \label{fig:div2k}
\end{figure}

To validate \ours, we design a controlled experiment to simulate various levels of geometric misalignment in a super-resolution (SR) scenario. This emulates real-world SR, in which the ground truth is imperfectly co-registered with the low resolution image due to sensor changes, motion, or delayed acquisition.
We use DIV2K \cite{div2k_Agustsson_2017_CVPR_Workshops} to create low/high resolution pairs by downsampling the high resolution images with $4\times$ bicubic interpolation.
Then, we simulate misalignments by introducing controlled translations and rotations into the pairs, abstracting away other factors such as sensor differences or radiometric variations.

\paragraph{Definition of spatial shifts}

To emulate realistic registration errors, we sample the shift magnitudes to favor small displacements but still produce larger ones occasionally.
The translation vector is defined by $\Delta x = r \cos\theta,\, \Delta y = r \sin\theta$, with amplitude $r$ sampled from a half-normal distribution of variance $\sigma_r^2$, and orientation angle uniformly sampled $\theta \sim \mathcal{U}(0, 2\pi)$. The rotation angle $\tilde{\theta}$ is sampled from $\mathcal{N}(0,\sigma_{\tilde{\theta}})$. 
As shown in \cref{fig:div2k_alignment}, we generate 3 misaligned datasets from DIV2K with 3 levels of alignment which we refer to as: \emph{High} ($\sigma_r=10, \tilde{\sigma}_{theta}=5$), \emph{Medium} ($\sigma_r=20, \tilde{\sigma}_{theta}=10$), \emph{Low} ($\sigma_r=35, \tilde{\sigma}_\text{theta}=15$).

\paragraph{Alignment score} In this controlled experiment, we know exactly the amplitude of the geometric shift for each image pair, and use it directly as the alignment score. This is a best case scenario which allows us to validate the principle of \ours. We define the shift $d_{\text{shift}}$ and the alignment $\gamma$ between $x_0$ and $x_1$ images of size $s$ (details in App.B) as:
\begin{equation*}
\resizebox{\columnwidth}{!}{$
d_{\mathrm{shift}}(x_0,x_1)
=\sqrt{r^2+(s\tilde{\theta}/2)^2},
\gamma(x_0,x_1)
=1-\frac{d_{\mathrm{shift}}(x_0,x_1)}{d_{\max}}
$}
\end{equation*}

\subsection{Cross-sensor super-resolution}
Cross-sensor super-resolution aims to translate observations acquired by a low-resolution (LR) source sensor to high-resolution images (HR) of a target sensor. This task is known to be especially sensitive to geometric distorsions between the LR/HR pairs \cite{michel_revisiting_cross_sensor}. We evaluate \ours on BreizhSR \cite{Okabayashi_2024_CVPR}, a cross-sensor satellite image SR dataset that pairs LR Sentinel-2 image time series at a \SI{10}{\meter\per\pixel} resolution with HR SPOT-6 data pre-processed at a \SI{2.5}{\meter\per\pixel} resolution. Since image acquisition is de-synchronized, image pairs are inevitably impacted by temporal gaps with changes in landscape. Moreover, BreizhSR covers the region of Brittany in France that exhibits strong seasonal variability due to its high fraction of agricultural areas (80\%). It is also known for its variable weather, introducing further temporal gap in image pairs since Sentinel-2 images with cloud coverage above 5\% are discarded. BreizhSR provides dates and coordinates as metadata.
\begin{figure}[h]
\begin{center}
\centerline{\includegraphics[width=\columnwidth, trim=50 0 0 40]{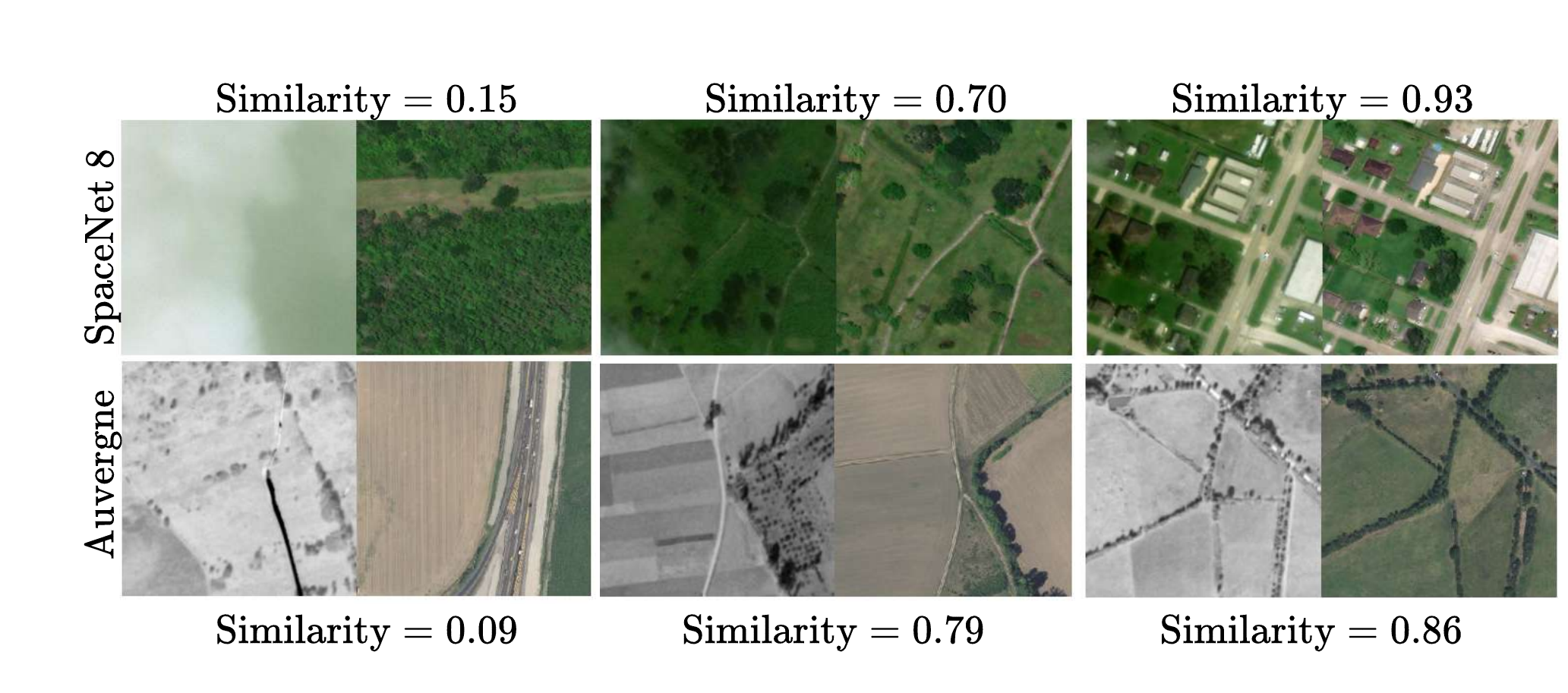}}
\vspace{-4mm}
\caption{Example pairs showing how DINOv3 embedding similarity, 
$S(E(x_0), E(x_1))$, reflects the shared semantic content between 
$x_0$ and $x_1$ on SpaceNet~8 and Auvergne datasets.}
\label{fig:emb_semantic}
\vspace{-12mm}
\end{center}
\end{figure}
\paragraph{Alignment score} In our experiments, we select the closest LR image to the HR image in time. We consider that discrepancies between paired images are primarily due to the acquisition time gap. Therefore, we use the temporal alignment score based on dates available in image metadata. We select $\Delta t_{\text{max}}=105$ days which corresponds to the largest temporal gap in the training data. 

\paragraph{Baselines} We compare \ours against two established SR models: ESRGAN \cite{esrgan} and SR3 \cite{sr3}. We adopt the ESRGAN architecture from Michel \etal \cite{michel_revisiting_cross_sensor} specifically for cross-satellite SR. We also compare \ours against two versions of Augmented BM: (1) trained all image pairs (\emph{AugBM}), and (2) trained only on ``aligned'' data (\emph{AugBM (filtered)}). For (2),  we kept images with temporal gaps of 1 day or less. This filtered baseline follows the common strategy in the cross-satellite SR literature to reduce changes in image LR/HR pairs \cite{sen2venus, aybar2025sen2naipv2}. Due to these inherent misalignments, it has been argued that weakly aligned datasets like BreizhSR \cite{Okabayashi_2024_CVPR} or WorldStrat \cite{cornebise2022open}, which pair real cross-sensor observations, are not suitable to train conventional SR models \cite{michel_revisiting_cross_sensr_sr, donike_trustworthy}. Yet, this filtering inevitably removes a large amount of usable data, as even imperfect pairs can still contain valuable spatial and contextual information.

\paragraph{Metrics} We consider commonly used SR metrics: Mean Absolute Error (MAE) and Root Mean Squared Error (RMSE) for pixel-level spectral fidelity; and three perceptual metrics: peak signal-to-noise ratio (PNSR), learned perceptual image patch similarity (LPIPS) \cite{zhang2018perceptual}, and structural similarity index measure (SSIM). 

\begin{figure*}[t]
\begin{center}
\centerline{\includegraphics[width=\textwidth, trim=0 0px 0px 0, clip]{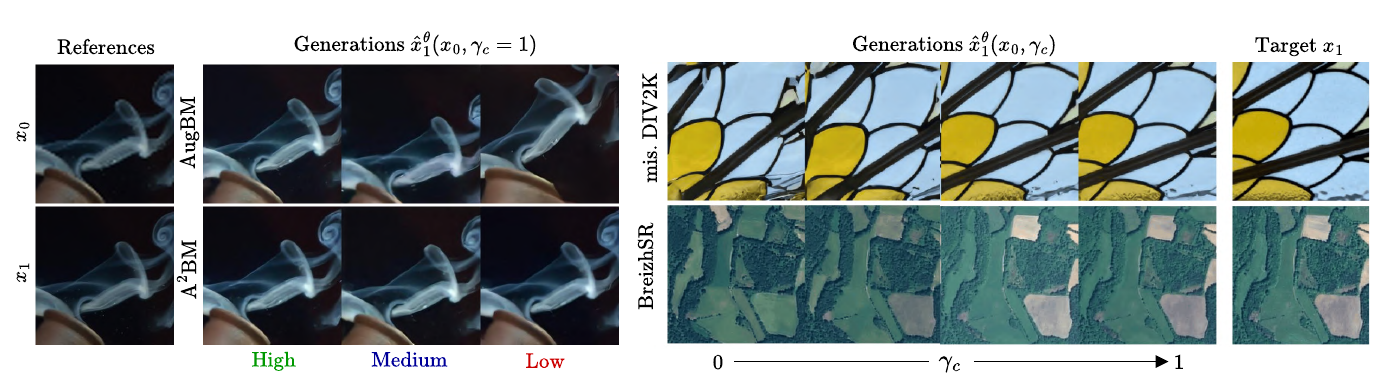}}
\caption{\textbf{(left)} Visual results on three modified versions of DIV2K with different alignement scores distributions (high/medium/low). AugBM distorts geometry, while \ours produces consistently aligned outputs on all three datasets, showing the effectiveness of the $\gamma_c$-conditioning. \textbf{(right)} Generated images from \ours with varying alignment scores ranging from $\gamma_c=0$ to $\gamma_c=1$. Top: synthetically misaligned DIV2K (\emph{Low} alignment), Bottom: BreizhSR. Generated images become more consistent with the target $x_1$ when $\gamma_c$ increases.}
\label{fig:gamma_impact}
\end{center}
\vspace{-10mm}
\end{figure*}

\subsection{Pixel-space domain adaptation}

We also evaluate \ours as an image translation approach for unsupervised domain adaptation (UDA). In this setup, we have access to a paired dataset with two domains $\mathcal{D}_0 = (\mathcal{X}_1, \emptyset)$ and $\mathcal{D}_1 = (\mathcal{X}_1, \mathcal{Y}_1)$, and to a \textit{predictive model} $S^1_\psi$ pretrained on $\mathcal{D}_1$. Domain adaptation aims to predict labels $y_0$ on images from $\mathcal{X}_0$ with $S_\psi^1$. We demonstrate the performance of \ours for UDA in pixel space, \ie, by translating $x_0$ to $x_1^{\text{gen}}$ and then predicting with $S_\psi^1$. In this scenario, we evaluate label-independent image translation methods to enable unsupervised domain adaptation, and we select \textit{semantic segmentation} as \textit{downstream task}. This allows us to quantify how much semantic information is preserved by the transport. We follow the protocol from \cite{le2026floweo} on two scenarios:

\noindent\textbf{1. Pre/post-disaster} domain adaptation on the SpaceNet~8 \cite{hansch_spacenet_2022} dataset, to address semantic shifts caused by flooding. SpaceNet~8 contains pairs of images acquired over the same area before and after a flood. The flood introduce significant semantic changes, such as clouds, turbulent floodwaters, and moving vehicles (e.g., boats on rivers) which degrade the alignment level.

\noindent\textbf{2. Historical-to-modern} on the Auvergne dataset \cite{lulc}, to bridge the gap caused by temporal and sensor shifts between historical (1946) and recent (2019) aerial imagery in the Auvergne area (France). In addition to the misalignments and artifacts due to the vastly different sensors employed, temporal evolution of the landscape between the two dates induces semantic changes: new buildings, new roads, cut down forests and crops lead to semantic inconsistencies. 

\paragraph{Alignment score}
In the pixel-space UDA setting, neither ground-truth labels nor auxiliary metadata are available to compute an alignment score directly. Recent applications of large pretrained visual encoders to semantic segmentation have demonstrated that their embeddings capture high-level semantic information \cite{pangaea}. Motivated by these findings, we employ an embedding-based alignment score using [CLS] tokens extracted from DINOv3 SAT-493M ViT-L \cite{simeoni2025dinov3} pretrained on remote-sensing imagery. The final alignment score is then computed according to \cref{eq:similarity_score}. As illustrated in \cref{fig:emb_semantic}, embedding similarity correlates well with semantic similarity between images.

\paragraph{Baselines} We compare \ours against several image translation baselines, ranging from adversarial-based approaches: Pix2pix \cite{pix2pix}, CycleGAN \cite{cyclegan_zhu}, StegoGAN \cite{wu2024stegogan}, to flow-based methods: UNSB \cite{kim2023unsb}, and FlowEO \cite{le2026floweo}. For fair comparison, we train all models with image pairs from the datasets, even for methods that generalize to unpaired translation (CycleGAN, StegoGAN, UNSB).

\paragraph{Metrics} We evaluate all models on \textit{downstream segmentation} metrics and on \textit{image quality} metrics. For \textbf{segmentation metrics}, we transfer the images $x_0$ from the test set of each dataset with the image translation models and then compute the prediction $\hat{y}_0 = S^1_\psi(x_1^\text{gen})$ with the segmentation model $S^1_\psi$ pretrained on $\mathcal{D}_1$. We compute \textit{mean Intersection over Union} (mIoU) and \textit{mean Accuracy} (mAcc). For \textbf{image quality}, we evaluate the fidelity of the generated images (perceptual similarity, absence of artifacts) with the \textit{Frechet Inception Distance} (FID) and the \textit{LPIPS} computed between translated images $x_1^\text{gen}$ and reference $x_1$ from the test set.

We also report the metrics obtained without any adaptation, and the upper-bound segmentation performance of the $S_1$ model when evaluated on the test set of $D_1$.

\subsection{\ours implementation details}
For cross-sensor super-resolution, we train \ours for \num{400}k training steps in pixel space. We use a \num{50}M parameters UNet backbone \cite{song_score-based_2020} and sampling is performed using an Euler sampler with 100 NFEs. For domain adaptation, we train \ours for \num{300}k training steps in the latent space of Stable Diffusion 3 (SD3) \cite{esser2024scaling}. During training, all images are encoded in SD3's latent space. At inference time, translation is provided in the latent space, and the generated latent is decoded to obtain the final image. We use a \num{230}M UNet backbone and an Euler sampler with 100 NFEs. In both cases, the alignment score is injected through a FiLM layer \cite{perez2018film} on top of the timestep encoding in the UNet.

%% file: sec/5_results.tex
\section{Results}
\label{sec:results}

\subsection{Synthetic experiments results}
\begin{table}[t]
    \centering
    \small
    \resizebox{\columnwidth}{!}{
        \begin{tabular}{l ccccc}
            \toprule
             & MAE$\downarrow$ & RMSE$\downarrow$ & SSIM$\uparrow$ & PSNR$\uparrow$ & LPIPS$\downarrow$ \\
            \midrule
            \rowcolor{gray!15} \multicolumn{6}{l}{\textit{Perfectly aligned} (upper-bound)} \\
            AugBM & 9.44 & 15.54 & 0.72 & 27.14 & 0.15 \\
            \midrule
            \rowcolor{customblue!50}\multicolumn{6}{l}{\textit{High: $\sigma_r=10, \sigma_\theta=5$}} \\
            AugBM & 24.17 & 38.12 & 0.40 & 18.44 & 0.28 \\
            \ours ($\gamma=1$) & \textbf{10.63} & \textbf{17.74} & \textbf{0.72} & \textbf{25.43} & \textbf{0.19} \\
            \rowcolor{customblue!50}\multicolumn{6}{l}{\textit{Medium: $\sigma_r=20, \sigma_\theta=10$}} \\
            AugBM & 28.30 & 43.02 & 0.38 & 17.10 & 0.36 \\
            \ours ($\gamma=1$) & \textbf{11.15} & \textbf{18.70} & \textbf{0.68} & \textbf{0.20} & \textbf{0.25} \\
            \rowcolor{customblue!50}\multicolumn{6}{l}{\textit{Low: $\sigma_r=35, \sigma_\theta=15$}} \\
            AugBM & 31.75 & 47.17 & 0.36 & 16.20 & 0.42 \\
            \ours ($\gamma=1$) & \textbf{13.41} & \textbf{22.28} & \textbf{0.61} & \textbf{23.16} & \textbf{0.21} \\
            \bottomrule
        \end{tabular}
    }
    \caption{\textbf{Quantitative results on the 3 DIV2K \cite{div2k_Agustsson_2017_CVPR_Workshops} variants with increasing misalignment levels.} (mean over 5 random seeds) \ours shows superior robustness as shift magnitude increases.}
    \label{tab:div2k_results}
    \vspace{-6mm}
\end{table}

Tab. \ref{tab:div2k_results} reports the quantitative results under decreasing levels of alignment, referred to as \emph{High}, \emph{Medium}, and \emph{Low} alignments.  We provide the upper bound performance when images are perfectly aligned i.e., the domain gap in pairs is reduced to bicubic downsampling. Quantitatively, the performance of AugBM deteriorates rapidly as the magnitude of the shifts increases. For instance, MAE rises from 9.44 to 24.17 and SSIM drops from 0.72 to 0.40 when adding small shifts (\emph{High} regime). In comparison, \ours maintains stable performance across all scenarios and remains close to the perfectly aligned upper-bound. Under the most challenging setting ($\sigma_r=35$, $\sigma_\theta=15$), \ours reduces the MAE by 51\% (13.41 vs. 31.75) and the RMSE by more than 52\% (22.28 vs. 47.17) relative to AugBM, while substantially improving perceptual quality (SSIM: 0.61 vs. 0.36, LPIPS: 0.21 vs. 0.42). Visual results in \cref{fig:gamma_impact} confirm that as dataset alignment degrades, AugBM fails to preserve structural fidelity during transfer, whereas \ours consistently maintains high-fidelity outputs. Overall, this indicates that \ours successfully leverages the alignment score and that is serves as an effective conditioning during inference when $\gamma_c=1$.

\subsection{Real-world datasets results}

\paragraph{Cross-sensor super-resolution} We report in \cref{tab:sr_weakly_results} the quantitative results on BreizhSR. First, we observe that bridge matching methods outperform ESRGAN and SR3, except for AugBM trained on the filtered dataset. 

SR3, which is diffusion-based, tends to hallucinate more than ESRGAN and bridge matching models, explaining its worse quantitative results despite producing visually appealing textures. Qualitative examples are provided in the supplementary material. \ours with $\gamma_c=1$ at inference achieves the best performance with a significant margin (+2\% in MAE and RMSE over AugBM). This shows the effectiveness of the alignment score conditioning. We further compare \ours to training AugBM on a filtered dataset where all image pairs with a temporal gap greater than one day were dropped, as per \cite{michel_revisiting_cross_sensr_sr, donike_trustworthy}. While filtering reduces misalignment, it also discards a lot of images, reducing diversity and therefore performance. In contrast, by explicitly informing the model of the alignment of the image pairs seen during training, \ours leverages the full dataset and achieves superior quantitative performance.

\begin{figure}[t]
    \centering
    \begin{subfigure}[t]{\linewidth}
        \centering
        \resizebox{\textwidth}{!}{
            \setlength{\tabcolsep}{0pt}
            \begin{tabular}{l cccc cccc}
        \toprule
        & \multicolumn{4}{c}{\textbf{SpaceNet 8}} & \multicolumn{4}{c}{\textbf{Auvergne}} \\
        \cmidrule(lr){2-5} \cmidrule(lr){6-9}
        & \multicolumn{4}{c}{\small Post-flood $\rightarrow$ Pre-flood} & \multicolumn{4}{c}{\small 1946 $\rightarrow$ 2019} \\
        \cmidrule(lr){2-5} \cmidrule(lr){6-9}
        \textbf{Method}
        & mIoU $\uparrow$ & mAcc $\uparrow$ & FID $\downarrow$ & LPIPS $\downarrow$
        & mIoU $\uparrow$ & mAcc $\uparrow$ & FID $\downarrow$ & LPIPS $\downarrow$ \\
        \midrule
        No adaptation   & 40.05 & 42.40 & 75.62 & 63.66 & 08.26& 21.82& 195.02& 72.00\\
        Upper bound     & 63.10 & 72.09 & 00.00 & 00.00 & 57.49& 68.75& 00.00  & 00.00 \\
        \midrule
        Pix2Pix   & 34.73 & 36.08 & 98.22          & 50.95& 16.03& 44.04& 84.96& 49.80\\
        CycleGAN  & 40.70 & 43.35 & 54.31& 55.70             & 22.16& 33.98& 52.67& 45.78\\
        UNSB      & 39.35 & 42.67 & 68.30          & 55.35             & \underline{33.30}& 47.94& 65.57& 46.81\\
        StegoGAN  & 38.62 & 40.58 & 66.61          & 58.07             & 22.30& 35.83& 60.28& 47.66\\
        FlowEO    & 44.65 & 48.79 & 60.32& 45.50& 16.98& 35.02& 101.93& 54.81\\
        \midrule
        AugBM & \underline{48.81}& \underline{54.51}& \underline{39.21}& \textbf{39.52}& 30.81& \underline{54.35}& \underline{13.08}& \underline{41.84}\\
        \rowcolor{customblue!50} \ours ($\gamma=1$)  & \textbf{49.75}& \textbf{54.91}& \textbf{25.00}& \underline{42.51}& \textbf{35.88}& \textbf{55.26}& \textbf{12.33}& \textbf{40.47}\\
        \bottomrule
    \end{tabular}
        }
        \caption{\textbf{Quantitative results} on domain adaptation datasets (mean over 5 random seeds). We report segmentation (mIoU, mAcc) and image quality metrics (FID, LPIPS).}
        \label{tab:da_results}
    \end{subfigure}
    \hfill
    \begin{subfigure}[t]{\linewidth}
        \centering
        \resizebox{\textwidth}{!}{
        \begin{tabular}{l ccccc}
        \toprule
        & \multicolumn{5}{c}{LowRes $\rightarrow$ HighRes} \\
        \cmidrule(lr){2-6}
        Method & MAE $\downarrow$ & RMSE $\downarrow$ & SSIM $\uparrow$ & PSNR $\uparrow$ & LPIPS $\downarrow$ \\
        \midrule
        ESRGAN & 14.06 & 20.31 & 0.58 & 22.66 & 0.22 \\
        SR3    & 35.87 & 46.69 & 0.41 & 15.27 & 0.38 \\
        \midrule       
        AugBM (filtered) & 14.35 & 20.80 & 0.57 & 22.70 & \textbf{0.19} \\
        AugBM & 13.17 & 19.63 & 0.62 & 23.17 & 0.23 \\
        \rowcolor{customblue!50} \ours ($\gamma=1$)        & \textbf{11.80} & \textbf{18.00} & \textbf{0.63} & \textbf{24.00} & 0.21 \\
        \bottomrule
    \end{tabular}
        }
        \caption{\textbf{Quantitative results} on the BreizhSR dataset (mean over 5 random seeds). We report reconstructions (MAE, RMSE) and perceptual (SSIM, PSNR, LPIPS) metrics.}
        \label{tab:sr_weakly_results}
    \end{subfigure}
    
    \vspace{-6pt}
    \captionof{table}{Quantitative results of \ours against various baselines on two challenging tasks: cross-sensor super-resolution on BreizhSR and domain adaptation for semantic segmentation on SpaceNet~8 and Auvergne datasets.}
    \vspace{-3mm}
\end{figure}

\begin{figure}[t]
    \centering
    \includegraphics[width=\linewidth, trim=5 0 15 0]{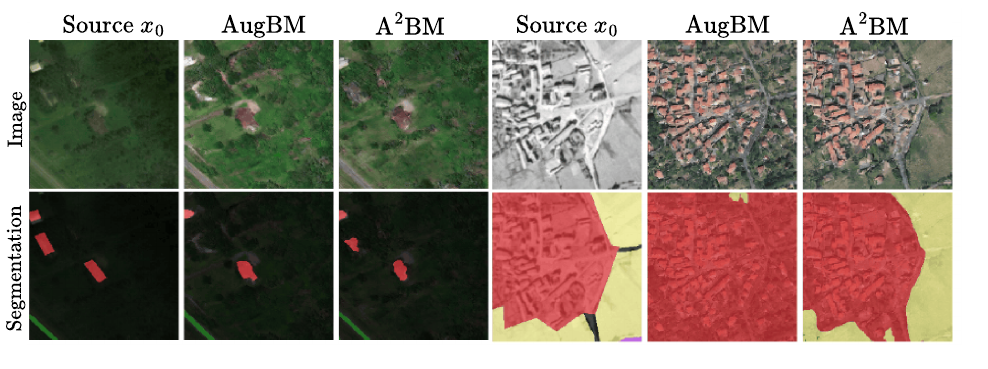}
    \vspace{-6mm}
    \caption{\textbf{Qualitative comparison of domain adaptation methods} on SpaceNet~8 (left) and Auvergne (right)
segmentation datasets. The first column shows the input image $x_0$; the second and third show images generated by AugBM and \ours. Below each image is the prediction from the segmentation model $\mathcal{S}_1$, and below $x_0$ the ground-truth mask $y_0$. \ours produces higher-fidelity images, yielding semantic maps closer to the ground truth due to alignment-score conditioning.
}
\vspace{-6mm}
    \label{fig:qualitative_da}
\end{figure}

\paragraph{Domain adaptation} We report in \cref{tab:da_results} the results in domain adaptation on the SpaceNet~8 and Auvergne datasets.  Results are averaged over 5 different inference runs. First, augmented bridge matching and \ours models consistently outperform baselines both in terms of downstream task and image quality metrics. On SpaceNet~8, AugBM and \ours demonstrate superior semantic preservation, achieving improvements of at least \num{4.16}\% in mIoU compared to the baselines. Moreover, on the Auvergne dataset, \ours outperforms the second best method (UNSB) by a large margin (+\num{2.58}\%).
More importantly, alignment conditioning consistently improves translation fidelity. We observe semantic segmentation gains on both datasets, with respective improvements of +\num{+0.94}\% and +\num{5.07}\% in mIoU on SpaceNet~8 and Auvergne. Qualitative results in \cref{fig:qualitative_da} show that, compared to AugBM, \ours yields higher-fidelity translations leading to 
improved building segmentation (2nd row). More qualitative results are included in Appendix A.2. Furthermore, 
\ours is less prone to hallucination: AugBM erroneously generates 
non-existing buildings in the Auvergne scene (2nd row).
While image quality is not our primary objective, \ours and AugBM also achieve the best image quality scores, both for FID and LPIPS. This shows that alignment conditioning not only enhances semantic fidelity but also increases generation quality.

\subsection{Alignment score impact}

We perform an ablation study to assess the influence of $\gamma_c$ at inference time.  \cref{fig:gamma_impact} illustrates the qualitative influence of the alignment conditioning parameter $\gamma_c$, on the synthetically misaligned DIV2K dataset and on BreizhSR. As $\gamma_c$ increases from 0 to 1, the generated outputs progressively aligns with the target image $x_1$. On the misaligned DIV2K dataset, the images exhibit geometric shifts with increasing rotation angles as $\gamma_c \rightarrow 0$. On BreizhSR, the agricultural crops on the image exhibit progressive color changes coherent with seasonal changes. Quantitatively, we find that metrics are positively correlated with increasing values of $\gamma_c$, \eg improving from an MAE of \num{30.07} ($\gamma_c=0$) to \num{11.80} ($\gamma_c=1$) on BreizhSR. We report more quantitative results in the appendix. This shows that the model has learned the different underlying alignment levels in the data. 

%% file: sec/6_conclusion.tex
\section{Conclusion}
\label{sec:conclusion}

We introduced \ours, a new image translation method that leverages semantic alignment between image pairs at training time to enhance fidelity to the source image at inference. By explicitly accounting for imperfectly aligned pairs, our method extends augmented bridge matching to realistic paired datasets where semantic correspondence is only approximate. \ours outperforms previous image translation baselines as well as augmented bridge matching on both synthetic and real-world datasets. We demonstrate the superiority of our approach on cross-sensor super-resolution and pixel-space unsupervised domain adaptation for semantic segmentation, two tasks of first importance for remote sensing. Finally, \ours paves the way for extending alignment conditioning to novel coupling strategies, enforcing translation fidelity on weakly paired and unpaired datasets.

%% file: supplemental.tex
\maketitle
\thispagestyle{empty}
\appendix

\section{Additional qualitative results}

\subsection{Qualitative results for cross-sensor super-resolution}

\begin{figure*}[h!]
\vskip 0.2in
    \begin{center}
    \centerline{\includegraphics[width=\linewidth, trim=10 0px 0px 0, clip]{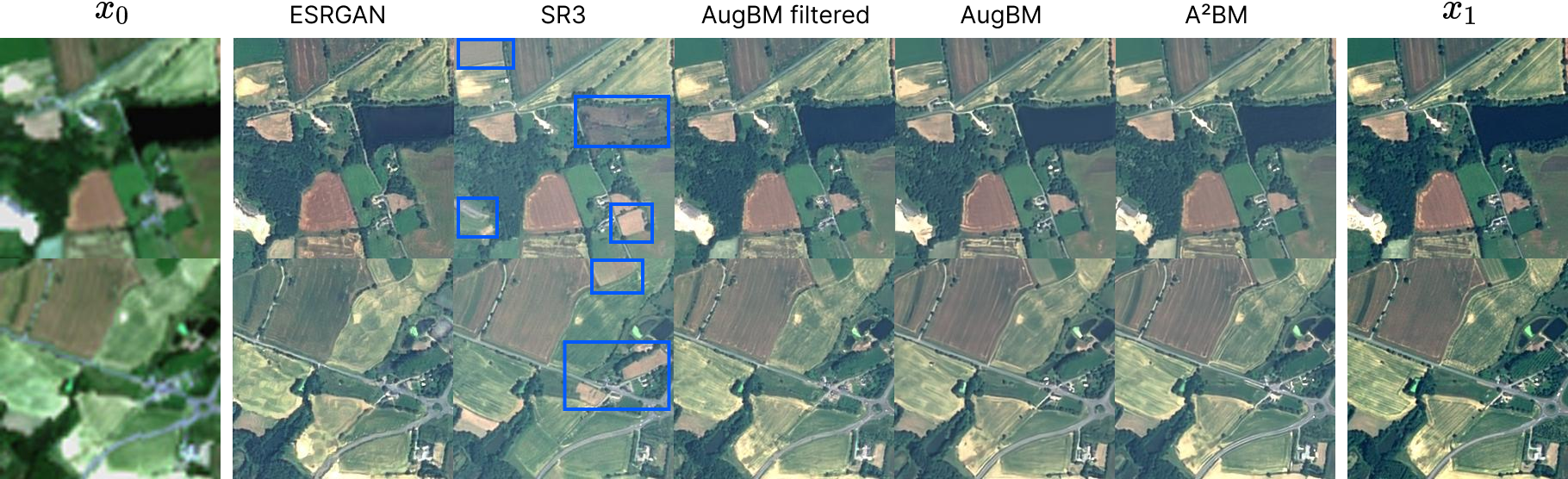}}
\caption{Qualitative results for cross-sensor super-resolution. From left to right: source image $x_0$, ESRGAN, SR3, AugBM trained on the filtered dataset (1 day), A$^2$BM, AugBM trained on the entire dataset, and the target image $x_1$. Hallucinations are highlighted in blue.}
    \label{fig:sr_visuals}
    \end{center}
\vspace{-10mm}
\end{figure*}

\cref{fig:sr_visuals} presents the visual results in cross-sensor super-resolution. SR3 tends to hallucinate details, which aligns with the observed quantitative results while still producing visually appealing outputs. ESRGAN produces blurred outputs compared to SR3 and bridge matching methods. AugBM filtered also fails to produces sharp textures, due to the training dataset size. A²BM preserves the visual quality of AugBM while achieving better quantitative performance.

\subsection{Qualitative results for domain adaptation}

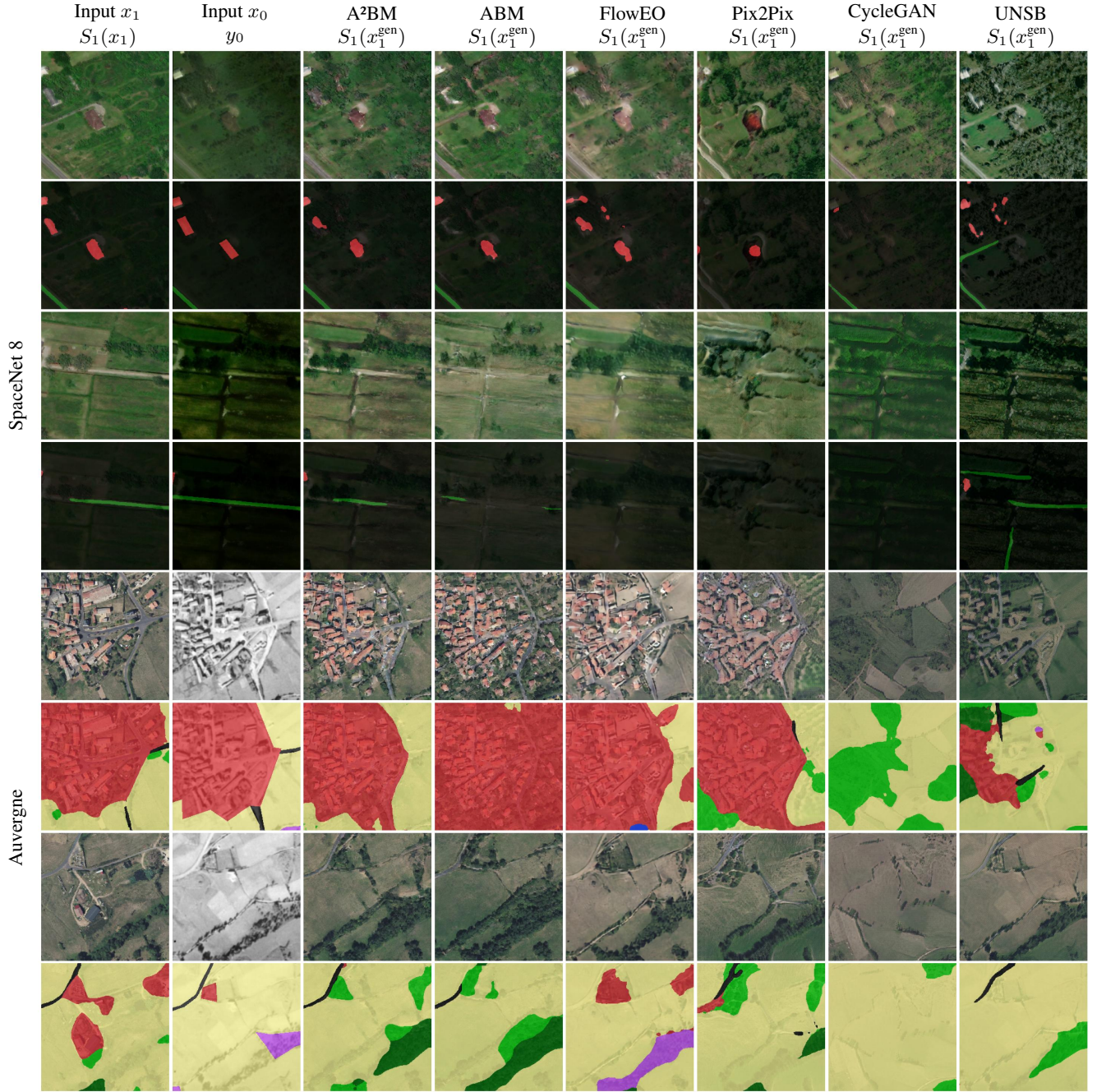
\begin{figure*}[h!]
\centering
\input{figures/visual_with_db}
\caption{Qualitative comparison of domain adaptation methods on segmentation datasets. The first column represents the dataset final image $x_1$, the second depict the initial image $x_0$, and the others display the images generated by the different methods. Below each image, we provide the corresponding prediction from the segmentation model $S_1$ or the true segmentation mask $y_0$ for the initial image (second column). \ours{} outperforms other methods in both semantic preservation (i.e., predicted mask closer to $y_0$) and image quality.}
\label{fig:visuals_beo}
\end{figure*}

In \cref{fig:visuals_beo}, we illustrate the data translation performance of \ours{} and compare the semantic masks predicted on the transferred images with those obtained by all baselines. This demonstrates that \ours{} maintains higher fidelity to the source image $x_0$ than competing baselines, resulting in improved domain adaptation performance for dense prediction tasks (here semantic segmentation).

\section{Ablation studies}

\subsection{Alignment score impact during inference}

In this section, we discuss the impact of the alignment score $\gamma_c$ on the metrics of downstream tasks.

\subsubsection{Super-resolution} 
\begin{figure}[h!]

     \centering
     \includegraphics[width=0.8\linewidth]{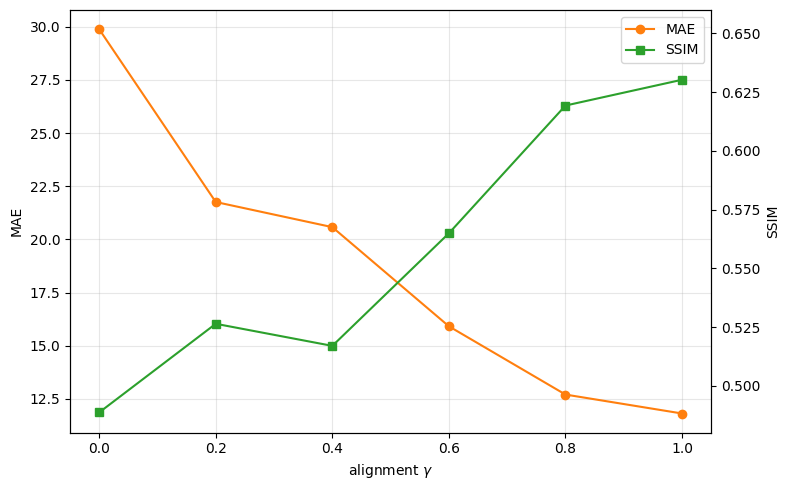}
     \label{fig:gamma_aligned}
     \caption{Quantitative influence of $\gamma_c$ at inference time for the cross-sensor super-resolution task. Metrics improve as $\gamma_c$ increases.}
\end{figure}

We assess the influence of $\gamma_c$ at inference time. \cref{fig:gamma_aligned} presents the evolution of MAE and SSIM as $\gamma_c$ increases ($\gamma_c \in \{0, 0.2, 0.4, 0.6, 0.8, 1\}$). The metrics improve as $\gamma_c$ increases, with MAE decreasing from $29.88$ ($\gamma_c=0$) to $11.84$ ($\gamma_c=1$) and SSIM increasing from $0.49$ to $0.63$. 


\subsubsection{Domain adaptation}

\begin{table}[h!]
    \resizebox{\linewidth}{!}{
    \setlength{\tabcolsep}{1pt}
    \centering
    \begin{tabular}{l lcccc cccc}
        \toprule
         && \multicolumn{4}{c}{\textbf{SpaceNet 8}} & \multicolumn{4}{c}{\textbf{Auvergne}} \\
        \cmidrule(lr){3-6} \cmidrule(lr){7-10}
         && \multicolumn{4}{c}{\small Post-flood $\rightarrow$ Pre-flood} & \multicolumn{4}{c}{\small 1946 $\rightarrow$ 2019} \\
        \cmidrule(lr){3-6} \cmidrule(lr){7-10}
        \textbf{Method}
         && mIoU $\uparrow$ & mAcc $\uparrow$ & FID $\downarrow$ & LPIPS $\downarrow$
        & mIoU $\uparrow$ & mAcc $\uparrow$ & FID $\downarrow$ & LPIPS $\downarrow$ \\
        \midrule
        \multicolumn{2}{c}{No adaptation}& 40.05 & 42.40 & 75.62 & 63.66 & 08.26& 21.82& 195.02& 72.00\\
        \multicolumn{2}{c}{Upper bound}& 63.10 & 72.09 & 00.00 & 00.00 & 57.49& 68.75& 00.00  & 00.00 \\
        \midrule
 \textbf{\ours}& $\gamma_c=0.0$& 49.28& \textbf{55.31}& 20.89& 40.10& 25.47& 45.73& 12.16&42.64\\
 & $\gamma_c=0.2$& 49.39& 55.26& \underline{20.77}& \underline{39.83}& 32.54& 54.74& \textbf{11.92}&41.95\\
 & $\gamma_c=0.4$& 49.36& \underline{55.29}& \textbf{20.65}& \textbf{39.78}& 26.12& 46.23& 12.19&42.26\\
 & $\gamma_c=0.6$& 49.57& 54.76& 21.64& 40.10& \underline{34.90}& \textbf{55.60}& 12.29&41.16\\
 & $\gamma_c=0.8$& \underline{49.74}& 54.97& 23.66& 41.30& 30.10& 47.11& \underline{12.12}&\underline{40.60}\\
        \rowcolor{customblue!50}&$\gamma_c=1.0$& \textbf{49.75}& 54.91& 25.00& 42.51& \textbf{35.88}& \underline{55.26}& 12.33& \textbf{40.47}\\
        \bottomrule
    \end{tabular}
    }
    \caption{\textbf{Ablation on alignment score}. We compare the results obtained with \ours for different values of $\gamma_c$. The mean intersection-over-union scores increase with the alignment score.}
    \label{tab:miou_gamma}
    
\end{table}

\subsection{Inluence of alignment score quality and bins}

\begin{figure}[h!]
    \centering
    \includegraphics[width=\linewidth]{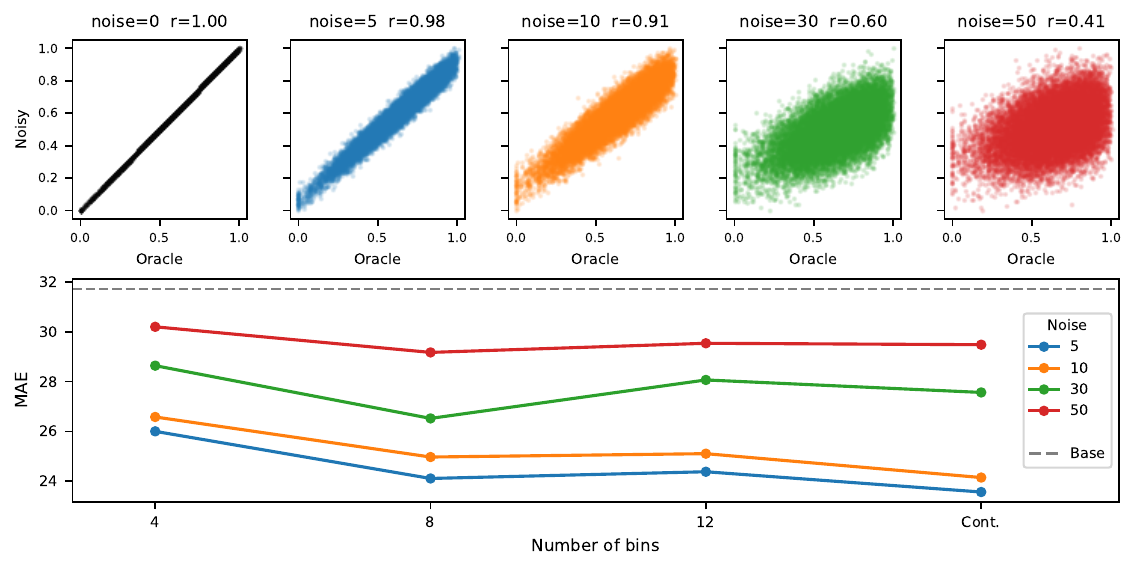}
    \vspace{-8.5mm}
    \caption{
\textbf{Top:} Oracle vs. noisy alignment scores for increasing noise levels, with Pearson correlation~$r$.
\textbf{Bottom:} Div2K MAE versus number of bins across noise levels (average over 5 seeds).
The gray line denotes AugBM baseline. Lower MAE is better.
}
    \label{fig:bins_ablation}
    \vspace{-2mm}
\end{figure}

We ablate the sensitivity to alignment score quality on DIV2K by perturbing the alignment scores with increasing noise levels, and the effect of bins in \cref{fig:bins_ablation}. The results show that conditioning on the alignment score remains beneficial even under strong noise.
 The trend suggests that finer conditioning (continuous score or large number of bins) is beneficial when the score is reliable. In contrast, binning appears more stable as score quality decreases, as binning removes unreliable fine-grained information.

\begin{figure}[t]
    \centering
    \includegraphics[width=\columnwidth]{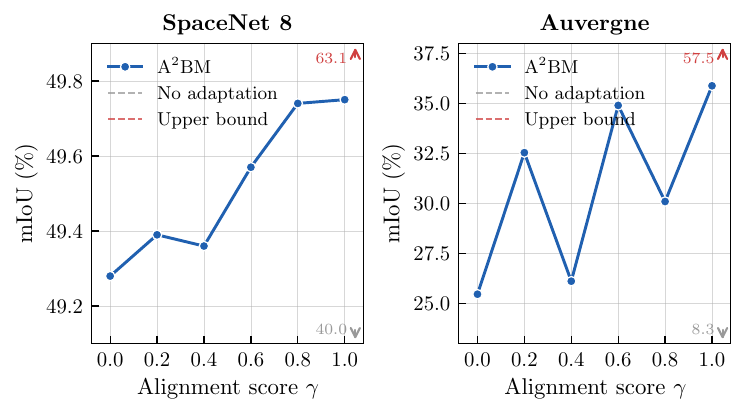}
    \caption{
        Ablation on the alignment score $\gamma_c$ for \ours on DA tasks.
        We report the mIoU on SpaceNet~8 and Auvergne datasets.
        On both datasets, mIoU generally increases with $\gamma_c$,
        confirming that \ours accurately learns trajectories guided
        by the alignment score.
    }
    \label{fig:gamma_da_visuals}
\end{figure}

We present in \cref{tab:miou_gamma} the evolution of the semantic segmentation metrics and generation quality metrics as $\gamma_c$ increases. We show in \cref{fig:gamma_da_visuals} that, on both datasets, the highest mIoU is obtained for the highest alignment score value, namely $\gamma_c=1$. The mIoU score generally increases with the alignment score. This highlights that \ours{} generates more faithful translated images when the alignment score is high.
Nevertheless, we do not observe any image quality improvements when increasing the alignment score, suggesting that the model remains a strong generator even when translating images with weaker (or no) conditioning on the initial image.

\subsection{Additional results for cross-sensor super-resolution}

For the filtered baseline, we initially discard image pairs with a time gap greater than 1 day, following common practices in cross-sensor super-resolution. This reduces the training set from \num{29920} pairs to \num{3176} pairs. We therefore report additional results using a less aggressive filtering strategy, where the temporal threshold is increased to 4 days, resulting in \num{16278} training pairs. We also report results for another strategy commonly used when dealing with weakly aligned data: weighted losses. We weight the loss by the alignment score associated with each training pair.


Relaxing the temporal filtering from $\leq 1$ day to $\leq 4$ days improves AugBM, reducing MAE from $14.35$ to $13.00$ and RMSE from $20.80$ to $19.34$. Weighting the loss based on alignment scores improves AugBM. However, \ours still achieves the best results with $11.80$ MAE and $18.00$ RMSE. Overall, we show that \ours is robust to different strategies for handling weakly aligned pairs.

\begin{table}[h!]
\centering
\resizebox{\linewidth}{!}{
\setlength{\tabcolsep}{1pt}
\begin{tabular}{llccccc}
\toprule
 &  & \multicolumn{5}{c}{LowRes $\rightarrow$ HighRes} \\
\cmidrule(lr){3-7}
Method & Setting & MAE $\downarrow$ & RMSE $\downarrow$ & SSIM $\uparrow$ & PSNR $\uparrow$ & LPIPS $\downarrow$ \\
\midrule
AugBM (filtered) & $\leq 1$ day & 14.35 & 20.80 & 0.57 & 22.70 & \textbf{0.19} \\
AugBM (filtered) & $\leq 4$ days & 13.00 & 19.34 & \textbf{0.63} & 23.50 & 0.23 \\
AugBM  & weighted loss & 12.80 & 19.09 & \textbf{0.63} & 23.39 & 0.22 \\
AugBM            & - & 13.17 & 19.63 & 0.62 & 23.17 & 0.23 \\
\rowcolor{customblue!50}
\ours ($\gamma_c=1$) & score conditioning & \textbf{11.80} & \textbf{18.00} & \textbf{0.63} & \textbf{24.00} & 0.21 \\
\bottomrule
\end{tabular}
}
\caption{\textbf{Additional quantitative results} on the BreizhSR super-resolution dataset. We compare different strategies for handling weakly aligned training pairs: strict temporal filtering ($\leq 1$ day), relaxed filtering ($\leq 4$ days), and loss weighted by alignment scores. \ours is trained on all pairs and achieves the best overall reconstruction performance.}
\label{tab:sr_additional_results}
\end{table}



\section{Dataset details}
\subsection{Synthetic experiments}
\label{sec:div2k_supp}

The alignment score is defined as the combination of the effects of translation and rotation applied to the image. The pixel displacement induced by translation is $r=\sqrt{\Delta x^2+\Delta y^2}$. The pixel shift induced by a rotation with an angle $\tilde{\theta}$ depends on its distance to the patch center.  For an image of size $s\times s$, we approximate the radius by $\frac{s}{2}$, and the global rotation shift as $\frac{s}{2}\tilde{\theta}$. Combining both effects, we define the global shift as: $d_{\text{shift}}(x_0,x_1)
\approx
\sqrt{
r^2 +
\left(\frac{s}{2}\tilde{\theta}\right)^2 }$. We then map this diplacement to an alignment score with $\gamma(x_0, x_1)=1-d_{\text{shift}}(x_0,x_1)/d_{\text{max}}$. We use  $d_{\text{max}}=45$ for \emph{High}, $d_{\text{max}}=85$ for \emph{Medium}, $d_{\text{max}}=120$ for \emph{Low} and clip values between \num{0} and \num{1}.

\subsection{Cross-sensor super-resolution}
The BreizhSR dataset contains Sentinel-2 low-resolution images paired with high-resolution SPOT-6 satellite imagery. Since the two sensors are not acquired simultaneously, temporal gaps between paired images may lead to landscape changes. The complete training set contains \num{29920} training image pairs and \num{13940} test pairs. We evaluate on aligned test pairs with a temporal gap of less than one day, resulting in \num{1189} pairs, following the common practice in cross-sensor super-resolution. 

\subsection{Domain adaptation}

For all datasets, we define three distinct splits: train, validation, and test. The training set is used to train both domain adaptation methods and predictive models. To reflect real-world scenarios -- where retraining a generative model on new data batches is impractical -- we restrict the training of image translation models to the training set. The validation set is used for hyperparameter tuning and model selection based on performance metrics, while the final reported metrics are computed on the test set.

\paragraph{SpaceNet 8:} \cite{hansch_spacenet_2022} is a segmentation dataset that contains pre and post-flood RGB images from Maxar for two different locations: Germany and Louisiana. The segmentation masks include three different classes: background, building, and roads. Original tiles are downsampled with a factor 2 and then cropped $256\times256$ images with an overlap of \num{70}\% to produce the training data. The final numbers of samples of each split are 22861/332/322.

\paragraph{Historical-to-modern:} the Auvergne dataset \cite{lulc} contains historical (1946) and recent (2019) aerial imagery in the Auvergne area (France). Original tiles are resampled to \num{1}m per pixel and then cropped $256\times256$ images with an overlap of \num{70}\% to produce the training data. The training set has 19991 images, and the validation and test sets have 63.

\section{Hyperparameters}
\label{sec:implem_details}

\subsection{Synthetic experiments}
We use a 50M-parameter UNet \cite{song_score-based_2020}. The model is trained for \num{300}k steps in pixel space with a learning rate of $2\times10^{-4}$, using gradient clipping and exponential moving averaging (EMA). The batch size is set 16.

\subsection{Cross-sensor super-resolution}
For a fair comparison, we use the same UNet backbone with 50M parameters for diffusion/BM methods, and increase the number of parameters of  Michel \etal's ESRGAN to \num{50}M parameters. We keep the same weights in the loss as the original implementation \cite{esrgan} which is also the ones used by Michel \etal. The generator was pretrained for \num{200}k iterations with using the $L_2$ loss, then the model was trained for \num{400}k iterations using the same learning rates and decay rates as the original paper. For SR3, we use T=500 timesteps for training and inference. SR3 and BM models were trained for \num{400}k iterations with a learning rate of 2$e^{-4}$ with gradient clipping and exponential moving average. Batch size is set to 16 in all experiments.

\subsection{Domain adaptation}

For domain adaptation, we train \ours for \num{300}k training steps in the latent space of Stable Diffusion 3 (SD3) \cite{esser2024scaling}. During training, all images are encoded in SD3's latent space. At inference time, translation is provided in the latent space, and the generated latent is decoded to obtain the final image. We use a \num{230}M UNet backbone from \cite{song_score-based_2020}, modified to integrate the alignment score through a FiLM layer \cite{perez2018film}.

We refer to the original articles and implementation for the baselines's hyperparameters.

\paragraph{FlowEO} We follow the implementation details from \cite{le2026floweo}. 
The latent space is from the distilled VAE \cite{bohan_madebyollin_2025}. We use a 120M parameters U-Net backbone and train it for \num{200000} steps using gradient clipping and exponential moving average. We use a learning rate of \num{1e-4} with \num{1000} steps of linear warmup and a batch size of 256.
At inference time, we set NFE=\num{50} and use an Euler ODE sampler. 

\paragraph{Pix2Pix} We use the reference PyTorch implementation available \footnote{\url{https://github.com/junyanz/pytorch-CycleGAN-and-pix2pix}}. Following the reference implementation, we train the models with a batch size of \num{1} for \num{200000} training steps with a learning rate of \num{2e-4} and learning rate linear decay. We use the \textit{LSGAN} \cite{mao2017least} adversarial loss. We tuned the $\lambda_\text{L1}$ parameter, decreasing it from \num{100} to \num{10} to improve image generation. We use a 9-blocks ResNet as a generator, and we use the PatchGAN discriminator with instance normalization.

\paragraph{CycleGAN} We use the same hyperparameters as for Pix2pix. We train the models with a batch size of \num{1} for \num{200000} training steps with a learning rate of \num{2e-4} and learning rate linear decay. Nevertheless, we do not decrease the value of $\lambda_\text{L1} = 100$ since we did not encounter any generation quality issue. We used the same network architectures as for Pix2Pix.

\paragraph{StegoGAN} We use the official implementation of StegoGAN \footnote{\url{https://github.com/sian-wusidi/StegoGAN}}.
We use \textit{LSGAN} adversarial loss, instance normalization, \num{200000} training iterations with a learning rate of \num{2e-4}. We set the same loss weight parameters as for the GoogleMismatch dataset in the original paper: $\lambda_A=10$, $\lambda_B=10$,$\lambda_A=10$, $\lambda_\text{id}=0.5$, $\lambda_\text{cycle}=0.5$ and $\lambda_\text{reg}=0.3$ for the mask regularization loss (similar to the original article values $\lambda_\text{cycle}=0.5$ for GoogleMismatch and $\lambda_\text{cycle}=0.3$ for PlanIGN).  As for Pix2Pix and CycleGAN, the generator is a 9-blocks-Resnet and we use the PatchGAN discriminator with instance normalization.

\paragraph{UNSB} We use the official implementation \footnote{\url{https://github.com/cyclomon/UNSB}} and compute \num{200000} training steps with a learning rate of \num{2e-4}. We use the proposed set of hyperparameters: $\lambda_\text{GAN} = 1$, $\lambda_\text{NCE} =1$, $\lambda_\text{SB} =1$. We leverage the same architectures as the other methods, namely 9-blocks-Resnet and PatchGAN discriminator with instance normalization.  Following original paper guidelines, we set the number of sampling steps to \num{5}.

\subsection{Computational cost}

Despite relying on SDE integration, \ours achieves domain adaptation in $\approx$\SI{128}{\milli\second}/image and pixel-space SR in $\approx$\SI{378}{\milli\second}/image (\cref{tab:computational_costs}). Diffusion and flow-based methods are inherently more computationally expensive than single-pass GANs, a trade-off widely accepted in the literature for improved training stability and generation quality. A$^2$BM introduces nearly no additional computational overhead compared to AugBM (see \cref{tab:computational_costs}).

\begin{table}[t]
    \centering
    \resizebox{\linewidth}{!}{%
    \setlength{\tabcolsep}{4pt}
    \begin{tabular}{l|c|c||l|c|c|c}
    \toprule
    \multicolumn{3}{c||}{\textbf{Domain Adaptation (A100 40GB)}} & \multicolumn{4}{c}{\textbf{Super-Resolution (A6000 48GB)}} \\
    \midrule
    Model & Inf. Time (s) & GFLOPs & Model& Param. & GFLOPs & Inf. Time (s) \\
    \midrule
    Pix2pix        & 0.03& 94.28& ESRGAN        & 51.87M & 447.61& 0.35\\
    CycleGAN       & 0.09& 337.12& SR3           & 54.01M & 249661.60 & 178.05\\
    StegoGAN       & 0.95& 722.79& - & -& -& -\\
    UNSB           & 0.11 (1)& 2861.31& - & -& -& -\\
    FlowEO         & 4.27& 1235.34&     -          &    -    &     -       &      -     \\
 AugBM& 15.577& 7971.88& AugBM         & 51.46M & 49932.37&36.04\\
    \rowcolor{customblue!50}A$^2$BM& 16.34& 7973.21 &               A$^2$BM&        51.54M &            49943.91&           38.11\\
    \bottomrule
    \end{tabular}
    }
    \vspace{-8pt}

    \caption{Computational costs. \textbf{Left:} Domain Adaptation on A100 40GB, batch size=128, except for UNSB (only supports  bs=1). \textbf{Right:} Super resolution on A6000 48GB, batch size=16. Flops are computed for a single image.}
    \label{tab:computational_costs}
    \vspace{-7mm}
\end{table}

\input{sec/proof_clean}




%% file: figures/visual_with_db.tex
\begin{tikzpicture}[scale=1, every node/.style={scale=1}]

        \node () at (0, 0) {
            \includegraphics[width=0.97\textwidth, trim=0 0 0 20, clip]{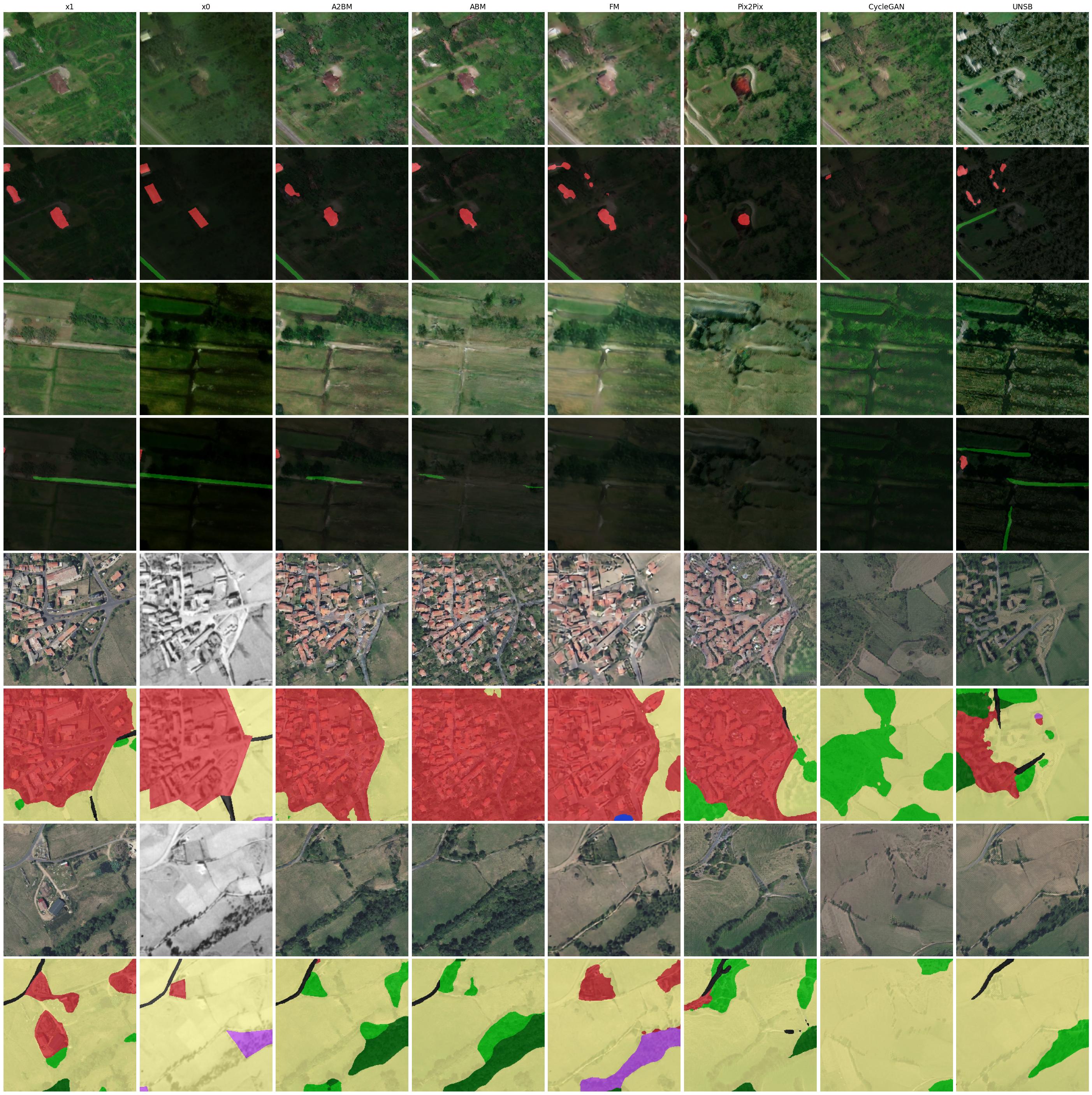}
        };
        \node () at (-7.35, 9) {\small{Input} $x_1$};
        \node () at (-7.35, 8.6) {\small{$S_1(x_1)$}};

        \node () at (-5.3, 9) {\small{Input} $x_0$};
        \node () at (-5.3, 8.6) {\small{$y_0$}};

        \node () at (-3.1, 9) {\small{\ours}};
        \node () at (-3.1, 8.6) {\small{$S_1(x^\text{gen}_1)$}};

        \node () at (-1, 9) {\small{\text{ABM}}};
        \node () at (-1, 8.6) {\small{$S_1(x^\text{gen}_1)$}};

        \node () at (1.1, 9) {\small{\text{FlowEO}}};
        \node () at (1.1, 8.6) {\small{$S_1(x^\text{gen}_1)$}};
        
        \node () at (3.2, 9) {\small{\text{Pix2Pix}}};
        \node () at (3.2, 8.6) {\small{$S_1(x^\text{gen}_1)$}};

        \node () at (5.28, 9) {\small{\text{CycleGAN}}};
        \node () at (5.28, 8.6) {\small{$S_1(x^\text{gen}_1)$}};

        \node () at (7.35, 9) {\small{\text{UNSB}}};
        \node () at (7.35, 8.6) {\small{$S_1(x^\text{gen}_1)$}};

        \node[rotate=90] () at (-8.8, 3) {\small{SpaceNet 8}};
        \node[rotate=90] () at (-8.8, -4.12) {\small{Auvergne}};
\end{tikzpicture}










%% file: sec/proof_clean.tex
\clearpage
\onecolumn
\section{\ours's velocity and drift derivations}
\label{sec:proof}
\subsection{General formulation}

In this section, we derive the drift function of A²BM. First, we consider a \textit{unconditional mixture of bridges} whose law over paths admits the decomposition:
\begin{equation}
    \mathbb{P}((x_t)_{t\in[0,1]})
    =
    \int_{\mathbb{R}^d \times \mathbb{R}^d}
    \mathbb{Q}\big((x_t)_t \mid x_0, x_1\big)
    \, \text{d}\mathbb{P}_{0,1}(x_0, x_1),
\label{eq:mixture}
\end{equation}
where $\mathbb{P}_{0,1}$ denotes a joint distribution over endpoints and $\mathbb{Q}(\cdot \mid x_0, x_1)$ is a scaled Brownian bridge conditioned on $(x_0, x_1)$. Capital $\mathbb{P}$ denotes a stochastic process (distribution over continuous paths $[0, 1] \rightarrow \mathbb{R}^d$) of law $p$. Intermediate marginals are denoted $p_t$ as previously and follow:
\begin{equation}
    p_t(x_t) = \int p_t(x_t|x_0,x_1)p(x_0,x_1)dx_0 dx_1,
\end{equation}
where the marginals of the $\sigma$-scaled Brownian bridge are given by $p_t(x_t|x_0,x_1) = \mathcal{N}((1-t)x_0 + tx_1, t(1-t)\sigma^2)$. In the case of paired dataset, we use \textit{data-dependent couplings}, i.e., $p(x_0, x_1) = p(x_1 \mid x_0)p(x_0)$. Thus, the marginals are given by:

\begin{equation}
    p_t(x_t) = \int p_t(x_t|x_0,x_1)p(x_1|x_0)p(x_0)\text{d}x_0 \text{d}x_1.
\label{proof:mixture}
\end{equation}

\paragraph{Latent variable $\gamma$:} 
To model varying degrees of alignment between $x_0$ and $x_1$, we enrich the definition of 
$p(x_1|x_0)$ with a latent variable $\gamma$. We assume $\gamma$ is independent of $x_0$, i.e. $p(\gamma|x_0)=p(\gamma)$, so that it acts as a global alignment control.
\begin{equation}
    p(x_1|x_0)
    = \int p(x_1|x_0, \gamma)\, p(\gamma)\, d\gamma.
\end{equation}

Thus, the marginals of the mixture of bridge ~\ref{eq:mixture} can be written as:
\begin{equation}
\begin{aligned}
p_t(x_t)
    &= \int p_t(x_t \mid x_0,x_1)
       \Bigg[\int p(x_1 \mid x_0, \gamma) p(\gamma) d\gamma \Bigg]p(x_0) \text{d}x_0 \text{d}x_1, \\
    &= \int  p_t(x_t|x_0,x_1)p(x_1 \mid x_0, \gamma)p(\gamma)p(x_0)\text{d}x_0 \text{d}\gamma \text{d}x_1.
\label{eq:inference}
\end{aligned}
\end{equation}

\subsection{Velocity and drift derivations}

\paragraph{Velocity}
Let the conditional marginal be defined as
\begin{equation}
p_t(x_t \mid x_0, \gamma)
=
\int p_t(x_t \mid x_0, x_1)\,p(x_1 \mid x_0, \gamma)\,dx_1,
\end{equation}
where $p_t(x_t \mid x_0,x_1)$ denotes the marginal of a $\sigma$-scaled Brownian bridge between $(x_0,x_1)$:
\begin{equation}
p_t(x_t \mid x_0,x_1)
=
\mathcal{N}\!\left((1-t)x_0 + tx_1,\; t(1-t)\sigma^2\right).
\end{equation}

Then the probability flow ODE
\begin{equation}
\dot x_t = u_t^\circ(x_t \mid x_0,\gamma)
\end{equation}
transporting $p_t(x_t \mid x_0,\gamma)$ admits velocity
\begin{equation}
u_t^\circ(x_t \mid x_0,\gamma)
=
\mathbb{E}_{p(x_1 \mid x_t,x_0,\gamma)}
\left[u_t(x_t \mid x_0,x_1)\right],
\end{equation}
where $u_t(x_t \mid x_0,x_1)$ denotes the velocity field of the Brownian bridge,
\begin{equation}
u_t(x_t \mid x_0,x_1)
=
\frac{1-2t}{2t(1-t)}\Big(x_t-(tx_1+(1-t)x_0)\Big) + (x_1-x_0).
\end{equation}

\paragraph{Drift}
\label{prop:drift}
Given this probability flow ODE with velocity $u_t^\circ(x_t \mid x_0, \gamma)$ generating $p_t(x_t \mid x_0, \gamma)$, there exists a SDE parameterized by a \textit{drift function} $b_t(x\mid x_0, \gamma)$ and a diffusion coefficient $\sigma \geq 0$:
\begin{equation}
    dx_t = b_t(x_t\mid x_0, \gamma)\,dt + \sigma dW_t,
\end{equation}

that generates the same marginals $p_t(x_t\mid x_0,\gamma)$ as the flow ODE, where $W_t$ is a standard Brownian motion.

In this case, the drift is given by
\begin{equation}
b_t(x_t \mid x_0,\gamma)
=
\frac{\mathbb{E}_{p(x_1 \mid x_t,x_0,\gamma)}[x_1] - x_t}{1-t}.
\end{equation}

\paragraph{Proof velocity}
First, we want to derive the \textit{velocity} $u^\circ_t$ of the probability flow ODE that describes the transport of the images from $p_0$ to $p_1$ conditioned on the initial $x_0$ and the alignment value $\gamma$:

\begin{align}
\frac{\partial p_t(x_t \mid x_0, \gamma)}{\partial t}
&= \frac{\partial}{\partial t}
\int p_t(x_t \mid x_0, x_1, \gamma)\, p(x_1 \mid x_0, \gamma)\, \text{d}x_1, \\
&= \int
\Bigg[\frac{\partial}{\partial t} p_t(x_t \mid x_0, x_1, \gamma)\Bigg]
p(x_1 \mid x_0, \gamma)\, \text{d}x_1.
\end{align}

The interpolant used $p_t(x_t|x_0,x_1) = \mathcal{N}((1-t)x_0 + tx_1, t(1-t)\sigma^2)$ is independent of $\gamma$. Then, $x_t \mid x_0, x_1$ is independent of $\gamma$ (note that $x_t$ depends on $\gamma$) and we have:
\begin{equation}
\frac{\partial p_t(x_t \mid x_0, \gamma)}{\partial t} = \int \Bigg[\frac{\partial}{\partial t}p_t(x_t \mid x_0, x_1)\Bigg]p(x_1 \mid x_0, \gamma)\,\text{d}x_1.
\end{equation}

As for $\gamma$-unconditional mixture of bridges, the $p(x_t \mid x_0, x_1)$ follows a continuity equation, and we denote $u_t$ its corresponding velocity:
\begin{equation}
\frac{\partial p_t(x_t \mid x_0, x_1)}{\partial t} = -\nabla_x \cdot \left[u_t(x_t \mid x_0, x_1)\,p_t(x_t \mid x_0, x_1)\right],
\end{equation}

where $u_t(x_t \mid x_0, x_1) = \frac{1 - 2t}{2t(1-t)}(x_t - (tx_1 + (1-t)x_0)) + (x_1 - x_0)$ \cite{tong_simulation-free_2023}.\\

Then, 

\begin{equation}
\begin{aligned}
\frac{\partial p_t(x_t \mid x_0, \gamma)}{\partial t}
&= - \int \nabla_x \cdot \left[u_t(x_t \mid x_0, x_1)\,p_t(x_t \mid x_0, x_1)\right]p(x_1 \mid x_0, \gamma)\,\text{d}x_1 \\
&= - \nabla_x \cdot \int u_t(x_t \mid x_0, x_1)\,p_t(x_t \mid x_0, x_1)\,p(x_1 \mid x_0, \gamma)\,\text{d}x_1 \\
&= - \nabla_x \cdot \int 
u_t(x_t \mid x_0, x_1)
\frac{p_t(x_t \mid x_0, x_1)p_t(x_t\mid x_0, \gamma)}{p_t(x_t\mid x_0, \gamma)}
\,p(x_1 \mid x_0, \gamma)\,\text{d}x_1 \\
&= - \nabla_x \cdot \int 
u_t(x_t \mid x_0, x_1)
\Bigg[
\frac{p_t(x_t \mid x_0, x_1)p(x_1 \mid x_0, \gamma)}
{p_t(x_t\mid x_0, \gamma)}
\Bigg]
\,p_t(x_t\mid x_0, \gamma)\,\text{d}x_1 \\
&= - \nabla_x \cdot 
\Bigg[
\int 
u_t(x_t \mid x_0, x_1)
\frac{p_t(x_t \mid x_0, x_1)p(x_1 \mid x_0, \gamma)}
{p_t(x_t\mid x_0, \gamma)}
\text{d}x_1
\Bigg]
\,p_t(x_t\mid x_0, \gamma)
\end{aligned}
\end{equation}

Using:
\begin{equation}
    p(x_1\mid x_t, x_0, \gamma) = \frac{p(x_t\mid x_0,x_1)p(x_1\mid x_0, \gamma)}{p(x_t\mid x_0, \gamma)},
\end{equation}

We have:

\begin{align}
\frac{\partial p_t(x_t \mid x_0, \gamma)}{\partial t} 
&= - \nabla_x \cdot \Bigg[\int u_t(x_t \mid x_0, x_1)p(x_1\mid x_t, x_0, \gamma)\text{d}x_1\Bigg]\,p_t(x_t\mid x_0, \gamma)\,\\
& = - \nabla_x \cdot \mathbb{E}_{p(x_1\mid x_t, x_0, \gamma)}[u_t(x_t \mid x_0, x_1)]\,p_t(x_t\mid x_0, \gamma)\,
\end{align}

We recognize the continuity equation and by identification, we have:

\begin{equation}
u^\circ_t(x_t \mid x_0, \gamma)= \mathbb{E}_{p(x_1\mid x_t, x_0, \gamma)}[u_t(x_t \mid x_0, x_1)].
\end{equation}

\paragraph{Proof drift}

\paragraph{Drift decomposition.} We start from the relation between the probability flow ODE and the SDE that generates the same marginals. The drift of the SDE satisfies the identity
\begin{equation}
b_t(x_t \mid x_0, \gamma)
=u_t^\circ(x_t \mid x_0, \gamma) + \frac{\sigma^2}{2}\nabla \log p_t(x_t \mid x_0, \gamma),
\end{equation}
which follows from the equivalence between the Fokker–Planck equation and the continuity equation \cite{song_score-based_2020,tong_simulation-free_2023}.

To obtain an explicit expression for the drift, we therefore need to express the $\gamma$-conditional score $\nabla_x \log p_t(x\mid x_0,\gamma)$ in terms of $p_t(x\mid x_0,x_1)$.

Deriving the $\gamma$-conditional score function:
\begin{equation}
\begin{aligned}
\nabla \log p_t(x_t\mid x_0, \gamma)
&= \frac{\nabla p_t(x_t\mid x_0, \gamma)}{p_t(x_t\mid x_0, \gamma)} \\
&= \frac{1}{p_t(x_t\mid x_0, \gamma)}
\nabla \int p_t(x_t, x_1\mid x_0, \gamma)\,\text{d}x_1 \\
&= \frac{1}{p_t(x_t\mid x_0, \gamma)}
\int \nabla p_t(x_t, x_1\mid x_0, \gamma)\,\text{d}x_1 \\
&= \frac{1}{p_t(x_t\mid x_0, \gamma)}
\int p_t(x_t, x_1\mid x_0, \gamma)
\nabla \log p_t(x_t, x_1\mid x_0, \gamma)\,\text{d}x_1 \\
&= \int
\frac{p_t(x_t, x_1\mid x_0, \gamma)}
{p_t(x_t\mid x_0, \gamma)}
\nabla \log p_t(x_t, x_1\mid x_0, \gamma)\,\text{d}x_1 .
\end{aligned}
\end{equation}
We have
\begin{equation}
\begin{aligned}
\frac{p_t(x_t, x_1\mid x_0, \gamma)}
     {p_t(x_t\mid x_0, \gamma)}
&=
\frac{p_t(x_t, x_1, x_0, \gamma)}
     {p_t(x_t\mid x_0, \gamma)p(x_0, \gamma)} \\
&=
\frac{p_t(x_1\mid x_0, x_t, \gamma)
      p_t(x_t\mid x_0, \gamma)
      p(x_0, \gamma)}
     {p_t(x_t\mid x_0, \gamma)p(x_0, \gamma)} \\
&=p_t(x_1\mid x_t,x_0,\gamma).
\end{aligned}
\end{equation}

which implies

\begin{equation}
\nabla \log p_t(x_t\mid x_0, \gamma)
=
\int p_t(x_1\mid x_0, x_t, \gamma)
\nabla \log p_t(x_t, x_1\mid x_0, \gamma)\,\text{d}x_1 .
\label{eq:score}
\end{equation}

In addition:

\begin{equation} 
p(x_t, x_1 \mid x_0, \gamma) = p(x_t \mid x_1, x_0, \gamma) p(x_1 \mid x_0, \gamma)
\end{equation}

\begin{equation} 
\log p(x_t, x_1 \mid x_0, \gamma) = \log p(x_t \mid x_1, x_0, \gamma) + \log p(x_1 \mid x_0, \gamma)
\end{equation}
\begin{equation} 
\nabla \log p(x_t, x_1 \mid x_0, \gamma) = \nabla \log p(x_t \mid x_1, x_0, \gamma) + \underbrace{\nabla \log p(x_1 \mid x_0, \gamma)}_{= 0}
\end{equation}

The second term vanishes because $p(x_1 \mid x_0, \gamma)$ does not depend on $x_t$. Therefore:
\begin{equation} 
\nabla \log p(x_t, x_1 \mid x_0, \gamma) = \nabla \log p(x_t \mid x_0, x_1, \gamma).
\end{equation}

Thus, from \cref{eq:score}:
\begin{equation} 
\nabla \log p_t(x\mid x_0, \gamma) =  \mathbb{E}_{p_t(x_1\mid x_0, x,  \gamma)} \left[\nabla \log p_t(x \mid x_0, x_1, \gamma)\right]
\end{equation}
Given that $x \mid x_0, x_1$ is independent of $\gamma$, in other words the interpolant is independent of $\gamma$:

\begin{equation} 
\nabla \log p_t(x\mid x_0, \gamma) =  \mathbb{E}_{p_t(x_1\mid x_0, x,  \gamma)} \left[\nabla \log p_t(x \mid x_0, x_1)\right]
\end{equation}

Using the drift decomposition, we have
\begin{equation} b_t(x \mid x_0, \gamma) = \mathbb{E}_{p(x_1\mid x_t, x_0, \gamma)}[u_t(x_t \mid x_0, x_1)] + \frac{\sigma^2}{2}\mathbb{E}_{p_t(x_1\mid x_0, x,  \gamma)} \left[\nabla \log p_t(x_t \mid x_0, x_1)\right].
\end{equation}

We remind the formulas for the velocity and score functions for $\gamma$-unconditional diffusion bridges \cite{tong_simulation-free_2023}:

\begin{equation}
    u_t(x_t \mid x_0, x_1) = \frac{1 - 2t}{2t(1-t)}(x_t - (tx_1 + (1-t)x_0)) + (x_1 - x_0)
\end{equation}

\begin{equation}
    \nabla \log p_t(x_t \mid x_0, x_1) = \frac{tx_1 + (1-t)x_0 - x_t}{\sigma^2t(1-t)}
\end{equation}

Let $\mu_t = x_t - (tx_1 + (1-t)x_0)$, then:

\begin{equation}
\begin{aligned}
b_t(x_t \mid x_0, \gamma) &= \mathbb{E}_{p(x_1 \mid x_t, x_0, \gamma)}\left[ u_t(x_t \mid x_0, x_1) + \frac{\sigma^2}{2} \nabla \log p_t(x_t \mid x_0, x_1) \right] \\
&= \mathbb{E}_{p(x_1 \mid x_t, x_0, \gamma)}\left[ \frac{1 - 2t}{2t(1-t)}(x_t - \mu_t) + (x_1 - x_0) + \frac{\sigma^2}{2} \cdot \frac{\mu_t - x_t}{\sigma^2\, t(1-t)} \right] \\
&= \mathbb{E}_{p(x_1 \mid x_t, x_0, \gamma)}\left[ \frac{1 - 2t}{2t(1-t)}(x_t - \mu_t) + (x_1 - x_0) - \frac{x_t - \mu_t}{2t(1-t)} \right] \\
&= \mathbb{E}_{p(x_1 \mid x_t, x_0, \gamma)}\left[ \frac{(1 - 2t - 1)(x_t - \mu_t)}{2t(1-t)} + (x_1 - x_0) \right] \\
&= \mathbb{E}_{p(x_1 \mid x_t, x_0, \gamma)}\left[ \frac{\mu_t - x_t}{1-t} + (x_1 - x_0) \right] \\
&= \mathbb{E}_{p(x_1 \mid x_t, x_0, \gamma)}\left[ \frac{t x_1 + (1-t)x_0 - x_t + (1-t)(x_1 - x_0)}{1-t} \right] \\
&= \mathbb{E}_{p(x_1 \mid x_t, x_0, \gamma)}\left[ \frac{x_1 - x_t}{1-t} \right] \\
&= \frac{\mathbb{E}_{p(x_1 \mid x_t, x_0, \gamma)}[x_1] - x_t}{1-t}.
\end{aligned}
\end{equation}